\pdfoutput=1

\documentclass[11pt]{article}

\usepackage[preprint]{acl}

\usepackage{times}
\usepackage{latexsym}

\usepackage[T1]{fontenc}

\usepackage[utf8]{inputenc}

\usepackage{microtype}

\usepackage{inconsolata}

\usepackage{graphicx}

\usepackage{amsmath} 
\usepackage{multirow}
\usepackage{multicol}
\usepackage{booktabs}  % Add this in your preamble
\usepackage{tabularx}
\usepackage{geometry}
\usepackage{enumitem}
\usepackage{subcaption}
\usepackage{makecell}
\usepackage[most]{tcolorbox}
\usepackage{algorithm}
\usepackage{algpseudocode}  % Modern replacement for algorithmic

\usepackage{tikz}
\usetikzlibrary{shapes, arrows, positioning}

\usepackage{booktabs,multirow,makecell,array,graphicx}
\usepackage{booktabs,multirow,makecell,array,tabularx,seqsplit}
\newcolumntype{Y}{>{\raggedright\arraybackslash}X}

\newcolumntype{M}[1]{>{\centering\arraybackslash}m{#1}} % v-centered fixed width

\renewcommand{\arraystretch}{0.95}

\title{Investigating Length Bias and Robustness in LVLMs for Multiple-Choice Question Answering}

\author{Md. Atabuzzaman \qquad Hani Alomari \qquad Chris Thomas \\
Department of Computer Science\\
  Virginia Tech \\
  \texttt{\{atabuzzaman, hani, christhomas\}@vt.edu}\\}
\begin{document}
\maketitle

\begin{abstract}
  Large Vision-Language Models (LVLMs) demonstrate strong performance on multiple-choice question answering (MCQA), yet their reliability under biased conditions remains underexplored. We investigate length bias, the systematic preference for longer answer options, through controlled benchmarks spanning fine-grained visual domains. LVLMs consistently favor longer options even when visually inconsistent, with accuracy dropping 2--38\% when correct answers are shorter and increasing 1--70\% when longer. This bias persists across model scales. Through layer-wise mechanistic analysis, we reveal that length bias emerges from late-layer suppression of visual evidence in favor of language-side length priors. Blurry-image experiments show that bias intensifies when visual evidence weakens, while abstention tests reveal that LVLMs struggle to refuse incorrect predictions. We propose and evaluate multiple inference-time mitigation strategies; while these provide partial improvements, none achieves deployable bias mitigation without relying on oracle information unavailable at inference time. These results underscore that inference-time interventions are insufficient, pointing to the need for architectural solutions and bias-aware training.\footnote{\texttt{Code and Dataset: \url{https://github.com/Atabuzzaman/Length_bias_robustness}}}
\end{abstract}

%%%%%%%%%%%%%%%%%%%%%%%%%%%%%%%%%%%%%%%%%%%%%%%%%%%%%%%%%%

\section{Introduction}

\begin{figure}[!t]
  \centering
  %\vspace{-1em}
  \includegraphics[width=1\linewidth, height=1.8in]{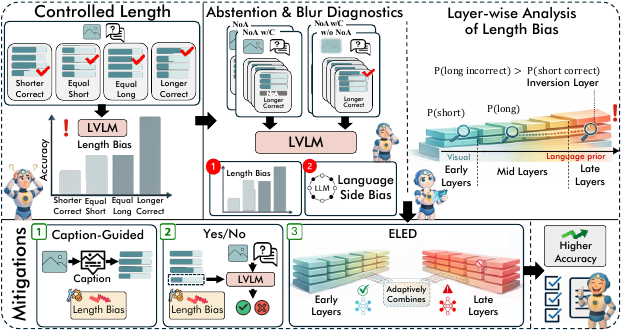}
  \vspace{-2em}
    \caption{\textbf{Overview of our framework} for investigating length bias in LVLMs. We create length-controlled MCQA variants to systematically expose length bias, conduct layer-wise analysis revealing late-layer suppression of visual evidence, perform diagnostic experiments using abstention and blurry images to localize bias sources, and evaluate mitigation strategies like ELED, caption-guided MCQA and Yes/No QA.}
   \label{fig:overview}
   \vspace{-2em}
\end{figure}

Large Vision-Language Models (LVLMs) \cite{llava2, llava, blip2, instructblip, molmo, Qwen2.5-VL, internvl2_5, gpt4o, januspro} have achieved impressive results across multimodal tasks including visual question answering (VQA), captioning, and fine-grained visual understanding \cite{finer, atabuzzaman2025}. However, recent evidence suggests these models exhibit systematic biases that compromise their reliability in structured decision-making tasks. In large language models (LLMs), biases in multiple-choice settings include preferences for specific answer positions, verbose options, and particular formatting patterns \cite{pezeshkpour2024large, zheng2023judging, zhao2025largelanguagemodelsbadly}. While early work suggests LVLMs may inherit or amplify such tendencies \cite{adila2024discovering, zong2024fool, atabuzzaman2025benchmarking}, the extent and mechanisms of these biases remain underexplored.

MCQA provides an ideal testbed for investigating these biases. By constraining outputs to a fixed set of options, MCQA enables controlled analysis of whether models base predictions on genuine visual-semantic understanding or rely on superficial heuristics. Recent work has identified arbitrary token and position biases in LVLM responses \cite{adila2024discovering, atabuzzaman2025benchmarking}, suggesting reliance on textual patterns rather than grounded reasoning. However, a comprehensive understanding of systematic preferences in LVLMs, particularly length bias and its interaction with visual reasoning, remains limited.

In this work, we conduct a systematic investigation of \textbf{length bias in LVLMs}: the tendency to prefer longer answer options regardless of visual evidence or correctness. Through carefully controlled benchmarks across five fine-grained visual domains, we demonstrate that LVLMs exhibit severe length bias, with accuracy dropping 2--38\% when correct answers are shorter and increasing 1--70\% when longer. We further conduct layer-wise probing analysis, revealing that while early layers encode sufficient visual evidence for correct answers, late decoder layers systematically suppress this evidence when longer options are present. This demonstrates that bias operates through late-layer override of visual evidence by language-side priors rather than initial visual encoding failures. This bias persists across model scales from 3B to 32B parameters and affects even state-of-the-art proprietary models, indicating deeply entrenched priors that extend beyond simple capacity limitations.

Beyond identifying length bias, we probe LVLM robustness under challenging conditions. To investigate whether bias originates primarily from visual processing or language-side reasoning, we conduct blurry-image experiments and find that length bias \textit{intensifies} under degraded visual input. This reveals that language-side heuristics become the dominant fallback when visual evidence is weak or ambiguous. Abstention tests using "None of the Above" options reveal poor abstention capability, with performance collapsing to 3--33.5\% when models should refuse to answer.

We propose and evaluate multiple mitigation strategies inspired by human reasoning, including (1) Early-Layer Ensemble Decoding (ELED), which combines predictions from early and late layers through adaptive decoding, (2) caption-guided answering with explicit visual grounding, (3) independent Yes/No evaluation of each option, and (4) selection by rejection, where models identify and discard incorrect options. While these approaches provide partial improvements, none achieves deployable bias mitigation in practice, revealing that inference-time interventions are insufficient as a practical solution. This suggests that architectural-level changes and bias-aware training are needed to fully address this bias.

\noindent
We make the following contributions:
\begin{itemize}[noitemsep, topsep=0pt]
\item We introduce controlled MCQA benchmarks across five fine-grained visual domains that systematically expose length bias in LVLMs.
\item We localize length bias to late decoder layers, demonstrating systematic override of visual evidence by language-side priors.
\item We conduct diagnostic analyses showing that bias intensifies under degraded visual input and LVLMs exhibit poor abstention capability.
\item We propose and evaluate multiple mitigation strategies (ELED, caption-guided MCQA, Yes/No Question-Answering, selection-by-rejection), revealing that inference-time methods are insufficient in practice.
\end{itemize}

%%%%%%%%%%%%%%%%%%%%%%%%%%%%%%%%%%%%%%%%%%%%%%%%%%%%%%%%%

\section{Related Work} 
Recent research has identified systematic biases in LLMs that affect their reliability in MCQA tasks, showing that models consistently favor certain tokens, positions, or options presented early in the sequence~\citep{zheng2023judging, pezeshkpour2024large, wang2025llms}. \citet{robinsonleveraging} showed that multiple-choice question (MCQ) presentation format influences model behavior, with performance varying significantly between prompts with explicit option identifiers versus cloze-style formats. Research suggests models struggle to effectively map option identifiers to their corresponding content~\citep{xue2024strengthened} or exhibit positional sensitivity combined with model uncertainty~\citep{pezeshkpour2024large}, with bias manifesting particularly strongly in knowledge-intensive scenarios~\citep{li2024can}. Mitigation strategies include neuron pruning~\citep{yang2024mitigating}, elimination of biased feed-forward network (FFN) components and attention heads~\citep{zhou2024unibias}, probability calibration~\citep{wei2024unveiling}, majority voting~\citep{gudatiny}, and causal debiasing~\citep{yang2025option}. \citet{ma2023poe} proposed a process-of-elimination approach that discards low-probability incorrect options before masking them to derive final predictions. Similarly, \citet{fu2025exclusion} introduced an exclusion-of-thought method combined with chain-of-thought prompting to reduce cognitive load through systematic option elimination in MCQA tasks. \citet{zhao2025largelanguagemodelsbadly} showed that LLMs are biased toward verbose options and that fine-tuning fails to mitigate this length bias in MCQA.

Recent work has explored bias mitigation in LVLMs through various approaches, including data augmentation~\citep{gokhale2020mutant}, model editing~\citep{cheng2023can, wang2024can}, and output post-processing~\citep{wang2021gender, zhang2024debiasing}. Approaches to address language priors include output probability calibration for classification tasks~\citep{zhang2024debiasing} and causal graph-based interventions on both visual and textual inputs for VQA~\citep{chen2024quantifying}. Researchers have explored activation-level interventions to reduce bias~\citep{adila2024discovering} and developed logit-based correction methods to address selection bias across varying task difficulty~\citep{atabuzzaman2025benchmarking}. However, no prior work has systematically examined length bias in LVLM-based MCQA through controlled experiments that isolate option length while maintaining visual accuracy. We address this gap through layer-wise mechanistic analysis, diagnostic experiments with degraded visual inputs and abstention tests, and evaluation of multiple mitigation strategies.

%%%%%%%%%%%%%%%%%%%%%%%%%%%%%%%%%%%%%%%%%%%%%%%%%%%%%%%%

\section{Benchmark Construction}
\label{sec:dataset}

\textbf{Base Dataset Collection.} We build our length-controlled MCQA benchmark using hard-difficulty questions from the fine-grained visual MCQA benchmarks of \citet{atabuzzaman2025benchmarking}. This dataset covers five fine-grained visual domains: CUB-200-2011 (200 bird species)~\citep{cub}, Stanford Dogs (120 dog breeds)~\citep{dogs}, FGVC Aircraft (102 aircraft variants)~\citep{aircrafts}, Stanford Cars (196 car models)~\citep{cars}, and Food-101 (101 food categories)~\citep{food}. Each base MCQ includes a question, four answer options representing different fine-grained classes within the same domain (e.g., different bird species or car models), and a single correct answer corresponding to the class depicted in the image.

%-------------------------------------------------------

\textbf{Length-Controlled Variant Generation.} To systematically investigate LVLMs' length bias, we use GPT-4o~\citep{gpt4o} to generate four length-controlled variants for each base MCQ while preserving semantic content and class structure. The base datasets already include two versions of each MCQ: one \textbf{with class names} explicitly mentioned (e.g., "This image shows a Yellow-headed Blackbird which has...") and one \textbf{without class names} that provides only visual descriptions (e.g., "This image shows a bird which has a black body, bright yellow head..."). We maintain both versions across all variants to analyze how bias changes with explicit class information. The with class name version examines whether domain knowledge can mitigate length bias, while the without class name version forces reliance on visual grounding, potentially increasing susceptibility to bias. We use carefully designed prompts to ensure consistency and validity across generated samples.

\noindent
We create the following four variants:

\textbf{Equal Long}: All four options have approximately the same length and contain detailed descriptions of the image class. This serves as a balanced baseline without length bias.

\textbf{Equal Short}: All options are short with approximately the same length (roughly half of Equal Long) and contain 2-3 visually identifiable attributes. This tests model performance with minimal descriptive content.

\textbf{Shorter Correct}: The correct answer is short (matching the Equal Short length) and includes 2-3 visually identifiable attributes that distinguish it from other options, while all three incorrect options are long (matching the Equal Long length). This configuration tests whether models exhibit bias against selecting shorter options.

\textbf{Longer Correct}: The correct answer is long (matching the Equal Long length) and includes multiple visually identifiable attributes that distinguish it from other options, while all three incorrect options are short (matching the Equal Short length). This configuration tests whether models exhibit bias toward selecting longer options.

To ensure high-quality variants that accurately test for length bias without introducing confounding factors, we manually inspect a sample of generated variants and enforce strict prompting constraints for semantic preservation, visual consistency, and structural uniformity (detailed in Appendix~\ref{dataset_constraints}). Additionally, for the Dogs dataset, we include samples from easy and hard difficulty categories to provide broader coverage of task complexity. \textbf{Our final datasets consist of 5,976 MCQs} across five domains with four length-controlled variants and two class-name settings per question.

%%%%%%%%%%%%%%%%%%%%%%%%%%%%%%%%%%%%%%%%%%%%%%%%%%%%%%%%%
\begin{figure*}[t]
  \centering
  
  \vspace{-1em}
  \includegraphics[width=1\linewidth, height=2.5in]{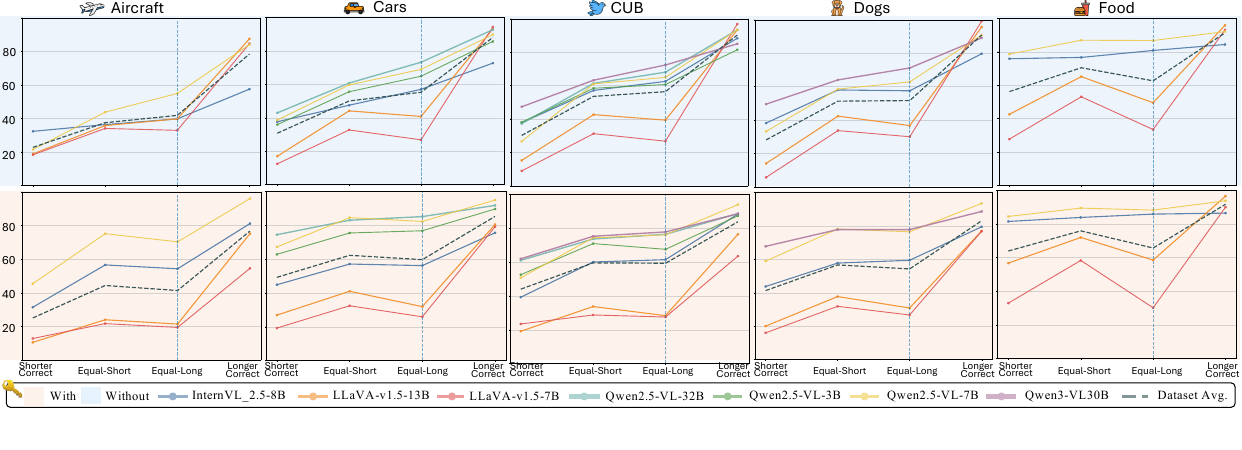}
   \vspace{-3.5em}
    \caption{\textbf{Accuracy (\%) of LVLMs across five datasets under different option-length conditions}, with and without class names. The vertical line at "Equal Long" indicates baseline performance for each LVLM-dataset pair.}
    \vspace{-0.2cm}
   \label{fig:table_1}
   \vspace{-1em}
\end{figure*} 
% the results figure
%%%%%%%%%%%%%%%%%%%%%%%%%%%%%%%%%%%%%%%%%%%%%%%%%%%%%%%%%
\section{Experimental Setup}
\label{sec:experimental_setup}

We evaluate length bias, robustness, and mitigation strategies using both open-source and proprietary LVLMs. Open-source models include Qwen3-VL-30B-A3B-Instruct~\cite{qwen3}, Qwen2.5-VL-3B-Instruct, 7B-Instruct, and 32B-Instruct (referred to as Qwen3-VL-30B, Qwen2.5-VL-3B, Qwen2.5-VL-7B, and Qwen2.5-VL-32B)~\cite{Qwen2.5-VL}, InternVL\_2.5-8B~\cite{internvl2_5}, LLaVA-v1.5-7B, and LLaVA-v1.5-13B~\cite{llava}. Proprietary models include OpenAI's GPT-4o, o4-mini (with reasoning)~\cite{gpt4o}, and Claude Sonnet-4.5 (used as C-Sonnet-4.5) (with reasoning)~\cite{sonnet45}. For each MCQ, we use 5 images per class for open-source models and 3 images per class for proprietary models due to API cost constraints. For example, in the CUB dataset (200 classes) with class names, we evaluate 1,000 images (200 × 5) per variant for open-source models and 600 images (200 × 3) for proprietary models. We apply this protocol across all experiments.

%%%%%%%%%%%%%%%%%%%%%%%%%%%%%%%%%%%%%%%%%%%%%%%%%%%%%%%%%
\begin{table}[!ht]
\centering
\begingroup
\setlength{\tabcolsep}{3.2pt}
\renewcommand{\arraystretch}{0.95}
\scriptsize
\resizebox{\columnwidth}{!}{%
\begin{tabular}{llccccc}
\toprule
\textbf{Dataset} & \textbf{Model} &
\textbf{\makecell{Equal\\Long}} & \textbf{\makecell{Equal\\Short}} &
\textbf{\makecell{Shorter\\Correct}} & \textbf{\makecell{Longer\\Correct}} \\
\midrule

\multirow{3}{*}{CUB (W/o)} 
  & GPT-4o & 71.17 & 62.83 & 45.17 & \textbf{92.83} \\
  & o4-mini & \textbf{71.67} & \textbf{66.00} & 44.33 & 91.00 \\ 
  & C-Sonnet-4.5 &65.33 &54.50	&\textbf{54.50} & 80.00 \\
\cmidrule(lr){1-6}

\multirow{3}{*}{Dogs (W/o)} 
  & GPT-4o & \textbf{86.11} & \textbf{75.56} & 55.83 & \textbf{97.78} \\
  & o4-mini & \textbf{86.11} & 73.61 & 46.94 & 96.67 \\ 
  & C-Sonnet-4.5 &68.06	&69.17	&\textbf{57.50}	&86.94 \\
\midrule

\multirow{3}{*}{CUB (W.)} 
  & GPT-4o &\textbf{ 89.50} & \textbf{86.33} & \textbf{80.33} & \textbf{95.83} \\
  & o4-mini & 81.67 & 81.33 & 73.33 & 89.83 \\ 
  & C-Sonnet-4.5 & 72.17 & 72.00 & 72.67 & 85.17 \\
\cmidrule(lr){1-6}

\multirow{2}{*}{Dogs (W.)} 
  & GPT-4o & \textbf{85.28} & \textbf{85.28} & \textbf{78.06} &\textbf{ 94.72} \\
  & o4-mini & 80.56 & 80.00 & 70.28 & 90.56 \\
  & C-Sonnet-4.5 & 68.06	& 70.83	 & 65.83 & 85.83 \\
\bottomrule
\end{tabular}%
}
\endgroup
\vspace{-0.25cm}
\caption{\textbf{Accuracy (\%) of advanced proprietary LVLMs} on CUB and Dogs datasets with (W.) and without (W/o) class names in the answer options.}
\vspace{-0.5cm}
\label{tab:gpt_model}
\end{table}

%%%%%%%%%%%%%%%%%%%%%%%%%%%%%%%%%%%%%%%%%%%%%%%%%%%%%%%%%

\section{Length Bias in LVLMs}
\label{length_bias}
To investigate the presence of systematic response preferences in LVLMs, we analyze how answer selection varies with the relative length of candidate options in MCQA settings. As shown in Figure~\ref{fig:table_1} (detailed results in Table~\ref{tab:model_length_bias_comparison}, Appendix~\ref{lvlm_performance}), all evaluated LVLMs exhibit measurable length bias, though its magnitude varies by model scale, dataset, and presence of class names within the options.

\noindent\textbf{LVLMs Overwhelmingly Prefer Longer Options, Even When Incorrect.} 
Across all LVLMs and datasets, models overwhelmingly prefer longer options, even when incorrect. This is most apparent in the \textit{Shorter Correct} condition, where accuracy drops sharply. For example, LLaVA-v1.5-13B drops by 24.10 percentage points (pp) on CUB (without class names) and 10.83\% on Dogs (with class names), while Qwen2.5-VL-7B suffers losses of 38.30\% on CUB and 15.10\% on Cars. Conversely, in the \textit{Longer Correct} setting, accuracy improves dramatically, often by 40--70\%, indicating reliance on textual length rather than visual evidence.

\noindent\textbf{Class Names Reduce but Do Not Eliminate Length Bias.} Including class names in candidate options reduces length bias for all LVLMs. On CUB, the performance drop in "Shorter Correct" reduces from 38.30\% to 25.40\% for Qwen2.5-VL-7B and from 24.10\% to 9.20\% for LLaVA-v1.5-13B when class names are included. However, length bias persists across all datasets even with class names, indicating that when visual distinctions are subtle, models rely on surface-level textual patterns rather than visual evidence.

\noindent\textbf{Scaling Models Improves Accuracy but Does Not Reduce Length Bias.} Comparing model families across parameter scales reveals that increasing model size consistently improves accuracy, particularly in Equal Long settings. On Cars (without class names), Qwen2.5-VL improves from 64.80\% (3B) to 73.16\% (32B) under "Equal Long" options. However, even the strongest models experience substantial drops when the correct option is shorter, and both LLaVA variants show 50--70\% performance boosts in "Longer Correct" conditions. This indicates scaling primarily enhances discriminative ability under balanced conditions but does not address reliance on option length as a heuristic.

\noindent\textbf{Proprietary and Advanced Models Demonstrate Reduced but Persistent Length Bias.}
Tables~\ref{tab:gpt_model} and~\ref{tab:model_length_bias_comparison} show that advanced models (GPT-4o, o4-mini, Claude Sonnet-4.5, Qwen3-VL-30B) substantially outperform smaller open-source models while exhibiting reduced length bias. With class names, GPT-4o achieves 89.50\% on CUB "Equal Long" compared to 76.10\% for Qwen2.5-VL-7B and maintains higher "Shorter Correct" accuracy (80.33\% vs 50.70\%). The gap between "Shorter Correct" and "Longer Correct" narrows significantly: GPT-4o shows 15.50 points on CUB and Claude Sonnet-4.5 shows 12.50 points, compared to 43.20 points for Qwen2.5-VL-7B. However, even these advanced models exhibit persistent length preference, with "Longer Correct" consistently outperforming "Shorter Correct," indicating that scaling and advanced training reduce but do not eliminate length bias.

\noindent\textbf{Length Bias Reflects a Learned Statistical Prior.}
The persistence of length bias across models, datasets, and scaling indicates it reflects a learned statistical prior rather than a decoding artifact. Option length functions as an implicit confidence cue that often outweighs visual evidence when distinctions are subtle, revealing that models rely on surface-level heuristics under uncertainty.

%%%%%%%%%%%%%%%%%%%%%%%%%%%%%%%%%%%%%%%%%%%%%%%%%%%%%%%%%%

\begin{figure}[!ht]
    \centering
    \includegraphics[width=\linewidth]{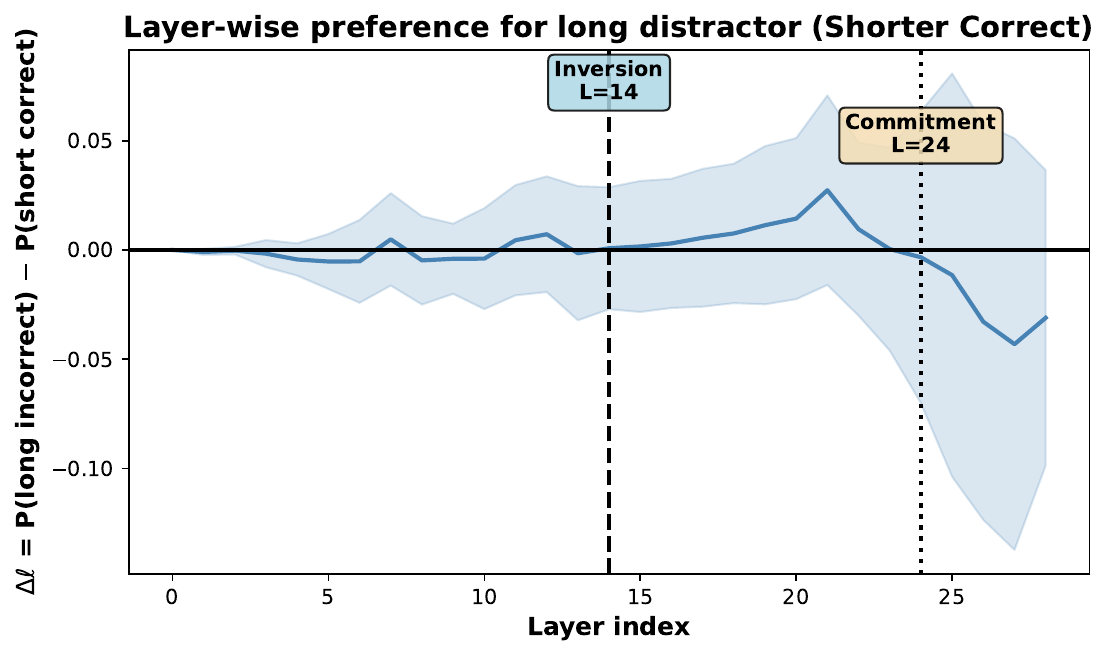}
    \vspace{-0.75cm}
    \caption{\textbf{Layer-wise preference for long distractors in the Shorter Correct setting using \textsc{Qwen2.5-VL-7B-Instruct}.} Mean preference $\Delta\ell = P(\text{long incorrect}) - P(\text{short correct})$ across 200 CUB examples via logit-lens decoding of hidden states (95\% CI shaded). Early layers show near-zero systematic preference for long options; mid layers exhibit a positive shift, indicating emergence of length bias in internal representations prior to decoding. Dashed line: \emph{stable inversion layer} ($L=14$), the first layer at which the mean $\Delta\ell$ exceeds zero for three consecutive layers. Dotted line: median \emph{commitment layer} ($L=24$), beyond which the predicted answer remains unchanged.
    }
    \vspace{-0.25cm}
    \label{fig:layerwise-delta-shorter-correct}
\end{figure}

\section{Where Length Bias Emerges in LVLM Computation}
\label{sec:layerwise_analysis}

To understand \emph{where} length bias arises inside LVLMs, we perform layer-wise analysis using logit-lens decoding: at each layer, we apply the language model head ($\mathrm{lm\_head}$) to the hidden state, restricted to the option-identifier tokens, to obtain a probability distribution over answer options in \textsc{Qwen2.5-VL-7B}. At each layer, we compute the model's probability assigned to a fixed long distractor (the longest of the three incorrect options) and to the short correct answer, tracking their difference $\Delta\ell = P(\text{long incorrect}) - P(\text{short correct})$.

As shown in Figure~\ref{fig:layerwise-delta-shorter-correct}, early layers exhibit near-zero or slightly negative values of $\Delta\ell$, indicating that initial representations do not favor longer options. Beginning around the stable inversion layer ($L=14$), the mean $\Delta\ell$ becomes consistently positive, revealing a \emph{semantic inversion} where long distractors overtake the short correct option within the model's internal computation on average.This transition occurs well before final decoding, demonstrating that length bias is introduced at the representation level rather than at final generation.

A representative individual example illustrates this progression at the single-example level: early layers favor the short correct option, but at layer 7, predictions first switch to a long distractor. By layer 11, the long distractor dominates consistently, and after layer 25, the model reaches a \emph{commitment point} where the predicted answer no longer changes. The final layer assigns substantially higher probability to the long distractor ($P=0.65$) than to the short correct option ($P=0.31$), demonstrating clear length-biased failure despite early evidence favoring the correct answer. Notably, $\Delta\ell$ decreases after the commitment layer, reflecting a shift to probability sharpening rather than option re-evaluation. Section~\ref{sec:conflict} analyzes this late-layer conflict resolution process in detail, demonstrating systematic override of visual evidence.

%%%%%%%%%%%%%%%%%%%%%%%%%%%%%%%%%%%%%%%%%%%%%%%%%%%%%%%%%

\begin{figure*}[!ht]
    \centering
    \includegraphics[width=0.9\linewidth]{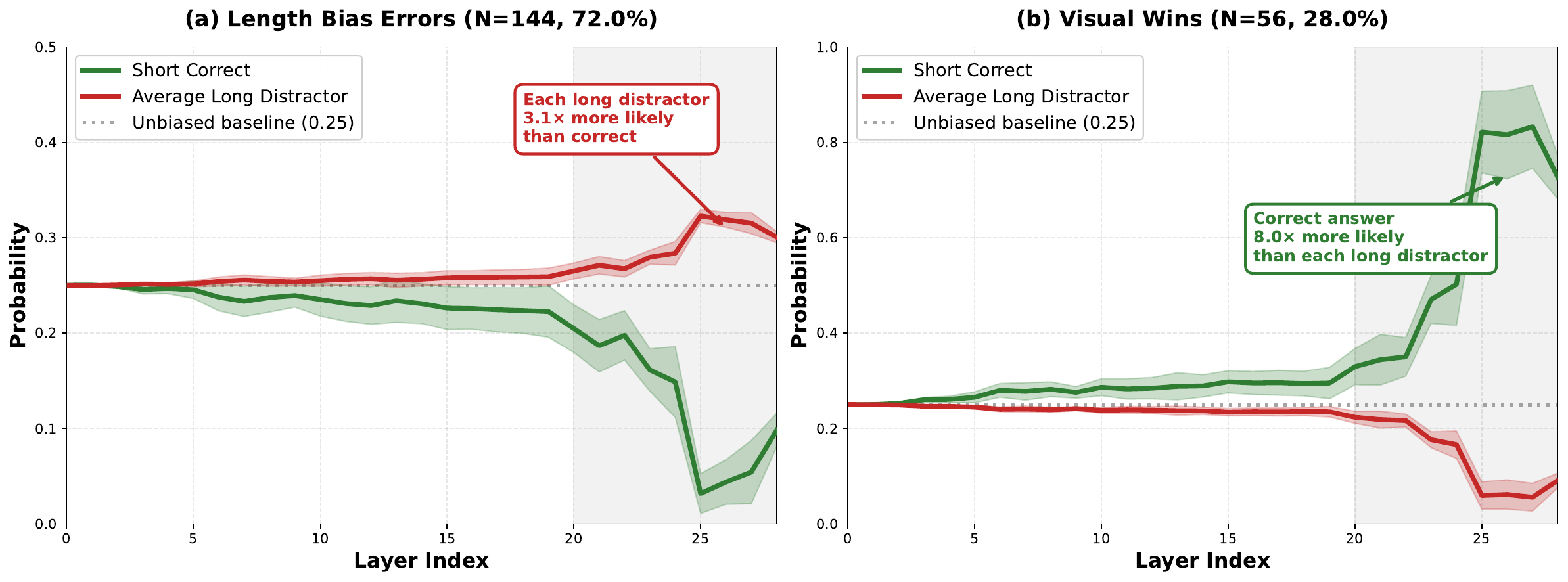}
    \vspace{-0.3cm}
    \caption{\textbf{Vision–language conflict resolution in late layers.} Layer-wise probabilities for the short correct answer (green) versus average probability of each long distractor (red), both starting at unbiased baseline 0.25 (dotted line). \textbf{(a)} In length-biased errors (N=144, 72\%), late layers (20--28, shaded) amplify long distractors to 0.31 while suppressing the correct answer to 0.10—each long option becomes $3.1\times$ more likely than correct. \textbf{(b)} When predictions align with ground truth (visual wins) (N=56, 28\%), the correct answer amplifies to 0.73 while each long distractor suppresses to 0.06 ($8.0\times$ preference ratio). Shaded regions: 95\% confidence intervals. Despite stronger amplification capacity for correct answers (+0.48 vs +0.06), the model utilizes this in only 28\% of cases.}
    \vspace{-0.3cm}
\label{fig:conflict}
\end{figure*}

\section{Vision–Language Conflict Resolution in Late Layers}
\label{sec:conflict}

To understand the mechanism underlying late-layer override of visual evidence, we analyze how competing signals evolve across all decoder layers. We use prediction correctness as an operational proxy for alignment with visual evidence. For 200 examples from the Shorter Correct split, we extract probabilities at each layer by applying the language model head to hidden states, tracking the short correct answer versus the average probability of each long distractor (mean of three long options). Both signals begin near the unbiased baseline of 0.25, enabling direct comparison of their evolution.

Figure~\ref{fig:conflict} reveals distinct conflict resolution patterns based on prediction outcomes. In length-biased errors (72\%, N=144), late layers (20--28) amplify each long distractor to 0.31 while suppressing the correct answer to 0.10—each long option becomes $3.1\times$ more likely than correct. When predictions align with ground truth (28\%, N=56), the pattern reverses dramatically: the correct answer amplifies to 0.73 while each long distractor suppresses to 0.06, yielding an $8.0\times$ preference ratio. Critically, the correct answer demonstrates stronger amplification capacity (+0.48 above baseline) than length bias (+0.06), yet this capacity is utilized in only 28\% of cases. This demonstrates that late layers perform winner-take-all conflict resolution but systematically fail to privilege signals aligned with visual evidence, revealing a failure mode where the same architectural components inconsistently resolve vision-language conflicts.

%%%%%%%%%%%%%%%%%%%%%%%%%%%%%%%%%%%%%%%%%%%%%%%%%%%%%%%%%

\begin{table}[!ht]
\centering
\small
\setlength{\tabcolsep}{4pt}
\renewcommand{\arraystretch}{0.96}
\resizebox{\columnwidth}{!}{%
\begin{tabular}{llccccccc}
\toprule
\textbf{Dataset} & \textbf{Setting} & \textbf{Model} &
\textbf{\makecell{Equal\\Long}} & \textbf{\makecell{Equal\\Short}} &
\textbf{\makecell{Shorter\\Correct}} & \textbf{\makecell{Longer\\Correct}} \\
\midrule

\multirow{4}{*}{CUB (W/o)} 
 & \multirow{2}{*}{w/o NoA} & Qwen2.5-VL-7B & 35.70 & 34.80 & \textbf{11.70} & 83.70 \\
 &  & InternVL\_2.5-8B & \textbf{32.10} & \textbf{31.90} & 21.90 & \textbf{74.30} \\ \cmidrule(lr){2-7}
 & \multirow{2}{*}{NoA w/C} & Qwen2.5-VL-7B & 25.20 & 18.60 & 11.60 & 66.40 \\
 &  & InternVL\_2.5-8B & \textbf{2.30} & \textbf{4.40} & \textbf{1.80} & \textbf{10.80} \\ 
\midrule

\multirow{4}{*}{Dogs (W/o)}
 & \multirow{2}{*}{w/o NoA} & Qwen2.5-VL-7B & 34.83 & \textbf{36.83} & \textbf{14.50} & 82.67 \\
 &  & InternVL\_2.5-8B & \textbf{32.50} & 40.00 & 26.17 & \textbf{76.67} \\ \cmidrule(lr){2-7}
 & \multirow{2}{*}{NoA w/C} & Qwen2.5-VL-7B & 24.00 & 18.50 & 11.33 & 62.67 \\
 &  & InternVL\_2.5-8B &\textbf{ 2.83} & \textbf{5.50} & \textbf{2.00} & \textbf{11.83} \\ 
\midrule

\multirow{4}{*}{CUB (W.)}
 & \multirow{2}{*}{w/o NoA} & Qwen2.5-VL-7B & 39.30 & 41.50 & \textbf{16.00} & 85.70 \\
 &  & InternVL\_2.5-8B & \textbf{30.90} & \textbf{34.60} & 18.20 & \textbf{80.10} \\ \cmidrule(lr){2-7}
 & \multirow{2}{*}{NoA w/C} & Qwen2.5-VL-7B & 20.50 & 17.00 & 10.00 & 60.80 \\
 &  & InternVL\_2.5-8B & \textbf{0.70} & \textbf{2.70} & \textbf{0.90} & \textbf{10.20} \\ 
\midrule

\multirow{4}{*}{Dogs (W.)}
 & \multirow{2}{*}{w/o NoA} & Qwen2.5-VL-7B & 35.83 & \textbf{36.83} & \textbf{16.50} & 80.50 \\
 &  & InternVL\_2.5-8B & \textbf{34.33} & 39.67 & 18.00 & \textbf{76.67} \\ \cmidrule(lr){2-7}
 & \multirow{2}{*}{NoA w/C} & Qwen2.5-VL-7B & 22.00 & 15.50 & 8.83 & 57.17 \\
 &  & InternVL\_2.5-8B & \textbf{3.00} & \textbf{4.67} & \textbf{1.67} & \textbf{12.67} \\
\bottomrule
\end{tabular}
}
\vspace{-0.25cm}
\caption{\textbf{Accuracy (\%) of LVLMs on blurry image evaluation} across CUB and Dogs datasets with (W.) and without (W/o) class names in options. "w/o NoA" denotes standard MCQ format without "None of the Above" options. "NoA w/C" includes "E. None of the above" while the correct answer remains present among A--D, testing over-abstention. For each setting, \textbf{bold} marks the \textbf{lowest} accuracy between the two models.}
\vspace{-0.75cm}
\label{tab:blurry_image}
\end{table}

%%%%%%%%%%%%%%%%%%%%%%%%%%%%%%%%%%%%%%%%%%%%%%%%%%%%%%%%

\section{Vision vs. Language Contributions to Length Bias}
\label{sec:vision-vs-language}

To causally test whether the conflict resolution failure in Section~\ref{sec:conflict} stems from insufficient visual information versus language-side heuristics, we degrade input images using Gaussian blur ($\sigma = 12$), reducing fine-grained details while preserving coarse structure. We test whether degraded visual input (1) intensifies length bias in standard MCQA, and (2) triggers appropriate abstention when a "None of the Above" option is available.

\noindent\textbf{Blurry Images Degrade Performance with Mixed Abstention Behavior.}
Table~\ref{tab:blurry_image} shows that blurred images substantially reduce accuracy and intensify length bias. For Qwen2.5-VL-7B on CUB without class names, performance drops from 64.90\% to 35.70\% in "Equal Long," and the gap between "Shorter Correct" (11.70\%) and "Longer Correct" (83.70\%) widens to 72 points versus 66.80 under clear images. Models show divergent abstention behavior: InternVL\_2.5-8B appropriately selects NoA at high rates (85--97\%), while Qwen2.5-VL-7B severely under-abstains (5--10\%), continuing to rely on length heuristics despite degraded visual evidence.

\noindent\textbf{Visual Degradation Accelerates Length Bias Emergence.}
Layer-wise analysis under visual degradation (Appendix~\ref{sec:effect_blur}) reveals that the median override layer (Figure~\ref{fig:blur}) shifts from $L = 19$ (clear images) to $L = 11$ (blurred images), demonstrating that stronger visual evidence delays—but does not prevent—bias emergence. Attention analysis (Appendix~\ref{sec:attention_analysis}) further shows that long incorrect options query visual features more than short correct answers, a difference partly attributable to their greater token length. Despite this increased visual attention, models still select options with visually inconsistent attributes, revealing systematic override of visual evidence. Blur accelerates this override by weakening visual evidence that contradicts incorrect attributes.

\noindent\textbf{Length Bias Is Language-Driven with Architecture-Dependent Robustness.}
The intensification of length bias under degraded visual input reveals that language-side heuristics dominate when visual evidence weakens. While length bias is fundamentally language-driven, abstention capability varies by architecture (InternVL: 85--97\% NoA; Qwen: 5--10\%), with implications for safe deployment where input quality cannot be guaranteed.

%%%%%%%%%%%%%%%%%%%%%%%%%%%%%%%%%%%%%%%%%%%%%%%%%%%%%%%%%%

\section{Mitigation Methods}
\label{sec:mitigation}

Having localized length bias to late decoder layers, we evaluate whether it can be mitigated at inference time, considering the following approaches.

%%%%------------------------------------

\subsection{Output-Level Calibration Baselines}
\label{sec:baselines}

We adapt two widely used calibration techniques from the LLM MCQA literature as baselines, and are the first to systematically evaluate them for length bias in LVLM-based MCQA. Both operate on the model's output distribution over options, without accessing intermediate representations.

\noindent\textbf{Pointwise Mutual Information (PMI)-based scoring.} Following \citet{holtzman2021surface}, we reweight each option by its unconditional likelihood to counter surface-form competition. For an option $o_k$, we score
\begin{equation}
\text{PMI}(o_k) = \log P(o_k \mid I, Q) - \log P(o_k),
\end{equation}
where $P(o_k \mid I, Q)$ is the conditional probability given image $I$ and question $Q$, and $P(o_k)$ is the option's prior probability estimated without visual grounding. This penalizes options that are intrinsically likely regardless of the image, partially offsetting the advantage of longer, more elaborate options.

\noindent\textbf{Length normalization.} We adjust conditional scores for token length by normalizing the log-probability of each option by its length raised to a power $\alpha$:
\begin{equation}
\text{score}(o_k) = \frac{\log P(o_k \mid I, Q)}{|o_k|^{\alpha}},
\end{equation}
where $|o_k|$ is the token length of option $o_k$ and $\alpha$ controls the strength of the correction. We report two settings, $\alpha=1.0$ (full normalization) and $\alpha=0.1$ (mild normalization), to characterize the method's sensitivity to this hyperparameter.

\begin{table}[t]
\centering
\small
\setlength{\tabcolsep}{4pt}
\renewcommand{\arraystretch}{0.96}
\resizebox{\columnwidth}{!}{%
\begin{tabular}{llccc}
\toprule
\multirow{2}{*}{\textbf{Dataset}} & \multirow{2}{*}{\textbf{Method}} & \textbf{Shorter} & \textbf{Longer} & \textbf{Avg.} \\
 & & \textbf{Correct} & \textbf{Correct} & \textbf{Gain} \\
\midrule
\multirow{4}{*}{CUB (W/o)}
 & Baseline            & 26.6 & 93.4 & -- \\
 & PMI                 & 46.5 & 74.5 & 0.5 \\
 & Len.\ Norm ($\alpha{=}1.0$) & 41.9 & 85.8 & \textbf{3.85} \\
 & Len.\ Norm ($\alpha{=}0.1$) & 30.5 & 92.2 & 1.35 \\
\midrule
\multirow{4}{*}{Dogs (W/o)}
 & Baseline            & 33.2 & 90.8 & -- \\
 & PMI                 & 48.5 & 73.5 & -1.0 \\
 & Len.\ Norm ($\alpha{=}1.0$) & 48.5 & 82.3 & \textbf{3.4} \\
 & Len.\ Norm ($\alpha{=}0.1$) & 36.3 & 90.0 & 1.15 \\
\midrule
\multirow{4}{*}{CUB (W.)}
 & Baseline            & 50.7 & 93.9 & -- \\
 & PMI                 & 69.1 & 75.7 & 0.1 \\
 & Len.\ Norm ($\alpha{=}1.0$) & 64.7 & 89.2 & \textbf{4.65 }\\
 & Len.\ Norm ($\alpha{=}0.1$) & 53.5 & 93.3 & 1.1 \\
\midrule
\multirow{4}{*}{Dogs (W.)}
 & Baseline            & 58.8 & 93.3 & -- \\
 & PMI                 & 69.7 & 78.5 & -1.95 \\
 & Len.\ Norm ($\alpha{=}1.0$) & 67.0 & 86.7 & \textbf{0.8 }\\
 & Len.\ Norm ($\alpha{=}0.1$) & 58.7 & 92.3 & -0.55 \\
\bottomrule
\end{tabular}%
}
\vspace{-0.75em}
\caption{Comparison of output-level calibration baselines on Qwen2.5-VL-7B across CUB and Dogs, with (W.) and without (W/o) class names. \textbf{Avg.\ Gain} is the mean change over the Baseline across the Shorter Correct and Longer Correct splits. Both PMI and length normalization improve Shorter Correct at the cost of degrading Longer Correct, yielding small or negative average gains.}
\label{tab:baseline-comparison}
\vspace{-0.45cm}
\end{table}

%%%%%%%%%%%%%%%%%%%%%%%%%%%%%%%%%%%%%%%%%%%%%%%%%

\begin{algorithm}[!ht]
\small
\caption{Split-Aware ELED}
\label{alg:eled}
\textbf{Require:} Hidden states $\{h_\ell\}_{\ell=0}^{L_{\max}}$, option texts $\{o_1,\dots,o_K\}$ with corresponding option-identifier token ids $\{t_1,\dots,t_K\}$, language model head $\mathrm{lm\_head}(\cdot)$, early layers $\mathcal{L}_{\text{early}}$, threshold $\tau = 1.2$, oracle split label $s \in \{\text{shorter correct}, \text{longer correct}, \text{equal}\}$

\textbf{Ensure:} Prediction $\hat{y}$
\begin{algorithmic}[1]
\State $z_{\text{early}} \gets \mathrm{lm\_head}\!\left(\frac{1}{|\mathcal{L}_{\text{early}}|}\sum_{\ell \in \mathcal{L}_{\text{early}}} h_\ell\right)$
\State $z_{\text{final}} \gets \mathrm{lm\_head}(h_{L_{\max}})$
\State $p_{\text{early}} \gets \mathrm{softmax}\big([z_{\text{early}}]_{t_1}, \dots, [z_{\text{early}}]_{t_K}\big)$
\State $p_{\text{final}} \gets \mathrm{softmax}\big([z_{\text{final}}]_{t_1}, \dots, [z_{\text{final}}]_{t_K}\big)$
\State $\hat{y}_{\text{early}} \gets \arg\max p_{\text{early}}, \quad \hat{y}_{\text{final}} \gets \arg\max p_{\text{final}}$
\If{$\hat{y}_{\text{early}} = \hat{y}_{\text{final}}$}
    \State \Return $\hat{y}_{\text{final}}$ \Comment{High confidence, no intervention}
\EndIf
\State $\ell_{\text{early}} \gets \text{length}(o_{\hat{y}_{\text{early}}}), \quad \ell_{\text{final}} \gets \text{length}(o_{\hat{y}_{\text{final}}})$
\If{$s = \text{shorter correct}$ \textbf{and} $\ell_{\text{final}} > \tau \cdot \ell_{\text{early}}$}
    \State \Return $\hat{y}_{\text{early}}$ \Comment{Length bias detected; correct toward short}
\ElsIf{$s = \text{longer correct}$ \textbf{and} $\ell_{\text{early}} > \tau \cdot \ell_{\text{final}}$}
    \State \Return $\hat{y}_{\text{early}}$ \Comment{Anti-length bias detected; correct toward long}
\Else
    \State \Return $\hat{y}_{\text{final}}$ \Comment{Default to mature representations}
\EndIf
\end{algorithmic}
\end{algorithm}

%%%-----------------------------------
\vspace{-0.1cm}
\subsection{Early-Layer Ensemble Decoding (ELED)}
\label{sec:eled}

Our layer-wise analysis reveals that length bias emerges progressively through the network rather than being encoded initially. For Qwen2.5-VL-7B, the mean $\Delta\ell$ trajectory reaches its stable inversion layer at $L=14$, with a median commitment layer of $L=24$ (Section~\ref{sec:layerwise_analysis}). For InternVL\_2.5-8B, bias remains near-zero through layer 16 before explosive growth from layers 17–28. This temporal separation suggests a natural mitigation: decode from earlier layers before bias dominates.

We first test whether decoding from pre-inversion layers can counteract length bias using a single fixed rule applied uniformly, with no information about which option length is correct. This rule averages logits over early layers (5--12 for Qwen2.5-VL-7B) and defers to the early-layer prediction whenever it disagrees with the final layer in a length-correlated way. Applied uniformly, this rule improves accuracy on Shorter Correct from 26.6\% to 45.5\% but substantially degrades Longer Correct from 93.4\% to 24.7\%: the same rule that overrides spuriously long distractors also overrides genuinely long correct answers, since early layers cannot distinguish the two without additional information. This mirrors the one-directional trade-off shown by the output-level calibration baselines (Table~\ref{tab:baseline-comparison}), motivating a split-aware gating mechanism that applies the correction selectively.

\noindent\textbf{Method.} Split-aware ELED averages logits across early layers and applies gating conditioned on an oracle split label $s$ indicating whether the correct answer is expected to be shorter or longer than the distractors (Algorithm~\ref{alg:eled}). For Qwen2.5-VL-7B, we use layers 5–12 (before the stable inversion layer at $L = 14$); for InternVL\_2.5-8B, we use layers 5–16 (before explosive onset at $L = 17$). Conditioning on $s$ allows the gate to apply the early-layer correction only in the direction indicated by the split, avoiding the one-directional trade-off observed above. Since $s$ is not available at inference time in deployment, we report these results as an oracle upper bound characterizing the ceiling achievable if the gating direction were known.

\begin{table}[!ht]
\centering
\begingroup
\setlength{\tabcolsep}{3.2pt}
\renewcommand{\arraystretch}{0.95}
\scriptsize
\resizebox{\columnwidth}{!}{%
\begin{tabular}{llccc}
\toprule
\textbf{Dataset} & \textbf{Model} & \textbf{\makecell{Shorter\\Correct}} & \textbf{\makecell{Longer\\Correct}} & \textbf{$\Delta$} \\
\midrule

\multirow{2}{*}{CUB (W/o)} 
  & Qwen2.5-VL-7B & 48.40 (+21.80)  & 95.00 (+1.60) & 46.60 (-20.20)  \\
  & InternVL\_2.5-8B & 52.10 (+14.30) & 90.40 (+2.10)  & 38.30 (-12.20)  \\ 
\cmidrule(lr){1-5}

\multirow{2}{*}{Dogs (W/o)} 
  & Qwen2.5-VL-7B & 46.49 (+13.32)  & 91.47 (+0.64) & 44.98 (-12.68)  \\
  & InternVL\_2.5-8B & 61.17 (+23.00) & 81.67 (+2.00) & 20.50 (-21.00) \\ 
\midrule

\multirow{2}{*}{CUB (W.)} 
  & Qwen2.5-VL-7B & 61.50 (+10.80) & 94.70 (+0.80) & 33.20 (-10.00)  \\
  & InternVL\_2.5-8B & 52.20 (+13.00)  & 90.40 (+2.10) & 38.20 (-10.90) \\ 
\cmidrule(lr){1-5}

\multirow{2}{*}{Dogs (W.)} 
  & Qwen2.5-VL-7B & 64.05 (+5.22) & 94.31 (+0.98) & 30.26 (-4.24) \\
  & InternVL\_2.5-8B & 57.83 (+14.16) & 81.83 (+2.50) & 24.00 (-11.66) \\
\bottomrule
\end{tabular}%
}
\endgroup
\vspace{-0.25cm}
\caption{\textbf{Split-aware ELED performance} for length bias mitigation, using oracle split information $s$ (Algorithm~\ref{alg:eled}; see Section~\ref{sec:eled} for details). Values in parentheses show changes over baseline.}
\label{tab:eled_result}
\vspace{-0.5cm}
\end{table}

\noindent\textbf{Results.}
Table~\ref{tab:eled_result} presents split-aware ELED results on CUB and Stanford Dogs with and without class names. On \textit{Shorter Correct}, split-aware ELED achieves substantial improvements: +21.8\% (Qwen/CUB W/o), +14.3\% (InternVL/CUB W/o), +13.3\% (Qwen/Dogs W/o), and +23.0\% (InternVL/Dogs W/o). Critically, \textit{Longer Correct} performance also improves (+0.6--2.5\%): conditioning the gate on the oracle split label $s$ allows early layers to correct in the appropriate direction for each split, rather than applying a single rule uniformly (cf.\ the motivation analysis above). The overall bias gap ($\Delta$ = Longer $-$ Shorter) reduces substantially: from 66.8\% to 46.6\% on Qwen/CUB (30.2\% relative reduction) and from 50.5\% to 38.3\% on InternVL/CUB (24.2\% relative reduction). For Equal-length conditions (Equal Long, Equal Short), the gating rarely triggers since all options have similar lengths, maintaining baseline performance. As Table~\ref{tab:eled_result} reflects an oracle upper bound, these results characterize the best achievable outcome given accurate gating direction, rather than a deployable inference-time method.

\noindent\textbf{Validation of mechanistic insights.}
Split-aware ELED's effectiveness validates our layer-wise analysis findings. First, substantial improvements on \textit{Shorter Correct} confirm that early layers (5--12 for Qwen, 5--16 for InternVL) encode less biased representations, directly validating our logit-lens measurements showing near-zero $\Delta_\ell$ in these layers. Second, improvements emerging specifically when gating detects length-correlated disagreements demonstrate that the observed bias corresponds to functionally relevant decision changes, not measurement artifacts. Third, consistent gains across architectures indicate this is a general property of current LVLM training, not model-specific.

\noindent\textbf{Comparison with other mitigation strategies.}
Beyond ELED, we evaluate three additional strategies inspired by human reasoning: caption-guided MCQA (Appendix~\ref{sec:caption_guided_mcqa}), Yes/No question answering (Appendix~\ref{sec:yes_no_qa}), and MCQA with selection by rejection (Appendix~\ref{sec:selection_by_rejection}).

%%%%%%%%%%%%%%%%%%%%%%%%%%%%%%%%%%%%%%%%%%%%%%%%%%%%%%%%%%%%%

% \input{fig/figure_2_split1}

\begin{figure}[!ht]
  \centering
  %\vspace{-1em}
  \includegraphics[width=\linewidth]{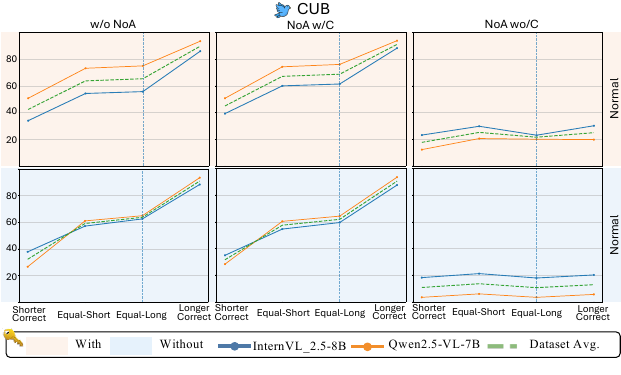}
   \vspace{-2em}
   \caption{Accuracy (\%) of LVLMs on "None of the Above" (NoA) evaluation settings across CUB datasets, reported both with and without class names. "w/o NoA" denotes the original MCQ format without a NoA option. "NoA w/C" appends a fifth option, "E. None of the above", while keeping the correct answer present among A–D (over-abstention test). "NoA wo/C" removes the correct answer and uses "D. None of the above." as the only valid choice (forced-abstention test).}
   \label{fig:table_2_noa_cub}
   \vspace{-1em}
\end{figure}

\section{Evaluating Abstention: When LVLMs Should Refuse to Answer}
\label{sec:noa_settings}

To probe LVLM robustness beyond length bias, we introduce a "None of the Above" (NoA) evaluation setting that tests models' ability to abstain when no valid option is present. This is essential for safe decision-making in high-stakes applications where overconfident guessing can be harmful. We design two complementary NoA configurations to distinguish between false abstention (unnecessarily choosing NoA) and failure to abstain (guessing when abstention is appropriate):

\textbf{Appended-NoA (Over-Abstention Test):}
We keep the original four options (A--D) unchanged and add a fifth option, "E. None of the above." The correct answer remains present among A--D. This evaluates whether models unnecessarily abstain when a valid answer is available, revealing tendencies toward excessive caution or uncertainty.

\textbf{Removed-Correct (Forced-Abstention Test):}
We remove the correct answer from the candidate set and shuffle the remaining incorrect options into positions A--C, while "D. None of the above" becomes the only valid response. This assesses whether models can correctly abstain when appropriate rather than guessing incorrectly.

%%%%%%%-------------------------------------------

\noindent\textbf{Results.}
We evaluate LVLM abstention capability under three settings. Figure \ref{fig:table_2_noa_cub} presents representative results; comprehensive results are provided in Appendix~\ref{NoA_results} (Figure~\ref{fig:table_2_noa} and Table~\ref{tab:NoA_comparison}).

\noindent\textbf{Models Exhibit Over-Abstention When NoA Is Added to Valid Options.}
When we append a NoA option while the correct answer remains available (NoA w/C), models show measurable over-abstention. For example, InternVL\_2.5-8B drops from 61.4\% to 55.7\% on CUB, indicating inappropriate selection of the abstention option.

\noindent\textbf{Models Struggle to Abstain When No Correct Answer Exists.}
When the correct answer is removed (NoA wo/C), performance collapses. Without class names, Qwen2.5-VL-7B falls to 3.2\% on CUB and 3.83\% on Dogs, indicating a near-total failure to select the abstention option even when it is the only valid answer, while InternVL\_2.5-8B achieves 17.7\% and 33.5\%, respectively. With class names, performance improves to 20.0\% and 11.33\% but remains far below acceptable levels, indicating models default to guessing even when semantic grounding cues are available.

\noindent\textbf{Models Handle Abstention Reliably Only in Trivial Cases.}
In Dogs (Easy), where distractors come from unrelated domains, both models achieve near-perfect abstention (100\%). This demonstrates that abstention failures are tied to fine-grained ambiguity rather than fundamental incapacity, revealing a critical robustness gap for safe deployment in challenging scenarios.

%%%%%%%%%%%%%%%%%%%%%%%%%%%%%%%%%%%%%%%%%%%%%%%%%%%%%%%%%%%%%%%%%

\section{Conclusion}
We present a systematic analysis of length bias and robustness in LVLMs for MCQA across five fine-grained visual domains. LVLMs consistently prefer longer answer options, with accuracy dropping substantially when correct answers are shorter and increasing dramatically when longer. This bias persists across model scales and state-of-the-art proprietary models. Through layer-wise mechanistic analysis, we reveal that length bias emerges from late-layer suppression of visual evidence in favor of language-side priors, where models systematically fail to privilege visual signals despite stronger amplification capacity for correct answers. Blurry-image experiments show that degraded visual input accelerates this override, confirming that bias stems from language-side heuristics dominating when visual evidence weakens. Abstention tests reveal poor capability, with performance collapsing when models should refuse to answer. We propose and evaluate multiple inference-time mitigation strategies; while these provide partial improvements, none achieves deployable mitigation without relying on oracle information. This underscores the need for architectural solutions and bias-aware training.

%%%%%%%%%%%%%%%%%%%%%%%%%%%%%%%%%%%%%%%%%%%%%%%%%%%%%%%%%

\section*{Limitations}
Our work focuses specifically on length bias and robustness in MCQA settings, not addressing other biases such as positional bias, modality imbalance, or cultural bias in LVLMs. Our benchmark is constructed from fine-grained visual classification MCQA across five domains; while this setting enables controlled isolation of length effects, direct empirical validation on broader VQA task families (e.g., open-ended VQA, compositional reasoning, or non-classification visual question answering) remains important future work, and the extent to which our findings generalize to these settings is an open question. The mitigation approaches we explore are inference-time strategies that do not modify model parameters. More effective solutions may require training-time methods such as bias-aware fine-tuning, including parameter-efficient approaches like Low-Rank Adaptation (LoRA), adapters, and prefix tuning, or architectural modifications that explicitly address length-based heuristics. Finally, our study examines a broad set of LVLM architectures spanning open-source models from 3B to 32B parameters, a mixture-of-experts model (Qwen3-VL-30B-A3B), and proprietary frontier models (GPT-4o, o4-mini, Claude Sonnet-4.5); continuous evaluation remains important as newer architectures continue to emerge.

%%%%%%%%%%%%%%%%%%%%%%%%%%%%%%%%%%%%%%%%%%%%%%%%%%%%%%%%%

\section*{Acknowledgments}
We acknowledge Advanced Research Computing (ARC) at Virginia Tech for providing the computational resources and technical support that contributed to the results reported in this paper. We thank the reviewers for their constructive feedback, which helped improve this paper.

%%%%%%%%%%%%%%%%%%%%%%%%%%%%%%%%%%%%%%%%%%%%%%%%%%%%%%%%%

\bibliography{custom}

@article{llava,
  title={Visual instruction tuning},
  author={Liu, Haotian and Li, Chunyuan and Wu, Qingyang and Lee, Yong Jae},
  journal={Advances in Neural Information Processing Systems},
  volume={36},
  year={2023}
}

@inproceedings{llava2,
  title={Improved baselines with visual instruction tuning},
  author={Liu, Haotian and Li, Chunyuan and Li, Yuheng and Lee, Yong Jae},
  booktitle={Proceedings of the IEEE/CVF Conference on Computer Vision and Pattern Recognition},
  pages={26296--26306},
  year={2024}
}

@article{instructblip,
      title={InstructBLIP: Towards General-purpose Vision-Language Models with Instruction Tuning}, 
      author={Wenliang Dai and Junnan Li and Dongxu Li and Anthony Meng Huat Tiong and Junqi Zhao and Weisheng Wang and Boyang Li and Pascale Fung and Steven Hoi},
      journal={Advances in Neural Information Processing Systems},
      volume = {37},
      year={2023}
}

@inproceedings{molmo,
  title={Molmo and {Pixmo}: Open weights and open data for state-of-the-art vision-language models},
  author={Deitke, Matt and Clark, Christopher and Lee, Sangho and Tripathi, Rohun and Yang, Yue and Park, Jae Sung and Salehi, Mohammadreza and Muennighoff, Niklas and Lo, Kyle and Soldaini, Luca and others},
  booktitle={2025 IEEE/CVF Conference on Computer Vision and Pattern Recognition (CVPR)},
  pages={91--104},
  year={2025},
  organization={IEEE}
}

@inproceedings{blip2,
  title={Blip-2: Bootstrapping language-image pre-training with frozen image encoders and large language models},
  author={Li, Junnan and Li, Dongxu and Savarese, Silvio and Hoi, Steven},
  booktitle={International conference on machine learning},
  pages={19730--19742},
  year={2023},
  organization={PMLR}
}

@misc{januspro,
      title={Janus-Pro: Unified Multimodal Understanding and Generation with Data and Model Scaling}, 
      author={Xiaokang Chen and Zhiyu Wu and Xingchao Liu and Zizheng Pan and Wen Liu and Zhenda Xie and Xingkai Yu and Chong Ruan},
      year={2025},
      eprint={2501.17811},
      archivePrefix={arXiv},
      primaryClass={cs.AI},
      url={https://arxiv.org/abs/2501.17811}, 
}

@misc{Qwen2.5-VL,
      title={Qwen2.5-VL Technical Report}, 
      author={Shuai Bai and Keqin Chen and Xuejing Liu and Jialin Wang and Wenbin Ge and Sibo Song and Kai Dang and Peng Wang and Shijie Wang and Jun Tang and Humen Zhong and Yuanzhi Zhu and Mingkun Yang and Zhaohai Li and Jianqiang Wan and Pengfei Wang and Wei Ding and Zheren Fu and Yiheng Xu and Jiabo Ye and Xi Zhang and Tianbao Xie and Zesen Cheng and Hang Zhang and Zhibo Yang and Haiyang Xu and Junyang Lin},
      year={2025},
      eprint={2502.13923},
      archivePrefix={arXiv},
      primaryClass={cs.CV},
      url={https://arxiv.org/abs/2502.13923}, 
}

@misc{internvl2_5,
      title={Expanding Performance Boundaries of Open-Source Multimodal Models with Model, Data, and Test-Time Scaling}, 
      author={Zhe Chen and Weiyun Wang and Yue Cao and Yangzhou Liu and Zhangwei Gao and Erfei Cui and Jinguo Zhu and Shenglong Ye and Hao Tian and Zhaoyang Liu and Lixin Gu and Xuehui Wang and Qingyun Li and Yiming Ren and Zixuan Chen and Jiapeng Luo and Jiahao Wang and Tan Jiang and Bo Wang and Conghui He and Botian Shi and Xingcheng Zhang and Han Lv and Yi Wang and Wenqi Shao and Pei Chu and Zhongying Tu and Tong He and Zhiyong Wu and Huipeng Deng and Jiaye Ge and Kai Chen and Kaipeng Zhang and Limin Wang and Min Dou and Lewei Lu and Xizhou Zhu and Tong Lu and Dahua Lin and Yu Qiao and Jifeng Dai and Wenhai Wang},
      year={2025},
      eprint={2412.05271},
      archivePrefix={arXiv},
      primaryClass={cs.CV},
      url={https://arxiv.org/abs/2412.05271}, 
}

@article{gpt4o,
      title={GPT-4o System Card}, 
      author={OpenAI},
      journal={arXiv preprint arXiv:2410.21276},
      year={2024}
}

@techreport{sonnet45,   
title={Claude Sonnet 4.5 System Card},   
author={Anthropic},  
institution={Anthropic},
year={2025},   
url={https://www.anthropic.com/claude-sonnet-4-5-system-card} 
}

@inproceedings{xue2024strengthened,
  title={Strengthened Symbol Binding Makes Large Language Models Reliable Multiple-Choice Selectors},
  author={Xue, Mengge and Hu, Zhenyu and Liu, Liqun and Liao, Kuo and Han, Honglin and Zhao, Meng and Yin, Chengguo and others},
  booktitle={Proceedings of the 62nd Annual Meeting of the Association for Computational Linguistics (Volume 1: Long Papers)},
  pages={4331--4344},
  year={2024}
}

@inproceedings{pezeshkpour2024large,
  title={Large Language Models Sensitivity to The Order of Options in Multiple-Choice Questions},
  author={Pezeshkpour, Pouya and Hruschka, Estevam},
  booktitle={Findings of the Association for Computational Linguistics: NAACL 2024},
  pages={2006--2017},
  year={2024}
}

@article{zheng2023judging,
  title={Judging llm-as-a-judge with mt-bench and chatbot arena},
  author={Zheng, Lianmin and Chiang, Wei-Lin and Sheng, Ying and Zhuang, Siyuan and Wu, Zhanghao and Zhuang, Yonghao and Lin, Zi and Li, Zhuohan and Li, Dacheng and Xing, Eric and others},
  journal={Advances in Neural Information Processing Systems},
  volume={36},
  pages={46595--46623},
  year={2023}
}

@inproceedings{robinsonleveraging,
  title={Leveraging Large Language Models for Multiple Choice Question Answering},
  author={Robinson, Joshua and Wingate, David},
  booktitle={The Eleventh International Conference on Learning Representations (ICLR)},
  year = {2023}
}

@inproceedings{li2024can,
  title={Can Multiple-choice Questions Really Be Useful in Detecting the Abilities of LLMs?},
  author={Li, Wangyue and Li, Liangzhi and Xiang, Tong and Liu, Xiao and Deng, Wei and Garcia, Noa},
  booktitle={Proceedings of the 2024 Joint International Conference on Computational Linguistics, Language Resources and Evaluation (LREC-COLING 2024)},
  pages={2819--2834},
  year={2024}
}

@inproceedings{zhou2024unibias,
  title={UniBias: Unveiling and Mitigating LLM Bias through Internal Attention and FFN Manipulation},
  author={Zhou, Hanzhang and Feng, Zijian and Zhu, Zixiao and Qian, Junlang and Mao, Kezhi},
  booktitle={The Thirty-eighth Annual Conference on Neural Information Processing Systems},
  year = {2024}
}

@inproceedings{yang2024mitigating,
  title={Mitigating Biases for Instruction-following Language Models via Bias Neurons Elimination},
  author={Yang, Nakyeong and Kang, Taegwan and Choi, Stanley Jungkyu and Lee, Honglak and Jung, Kyomin},
  booktitle={Proceedings of the 62nd Annual Meeting of the Association for Computational Linguistics (Volume 1: Long Papers)},
  pages={9061--9073},
  year={2024}
}

@inproceedings{wang2025llms,
  title={LLMs May Perform MCQA by Selecting the Least Incorrect Option},
  author={Wang, Haochun and Zhao, Sendong and Qiang, Zewen and Xi, Nuwa and Qin, Bing and Liu, Ting},
  booktitle={Proceedings of the 31st International Conference on Computational Linguistics},
  pages={5852--5862},
  year={2025}
}

@inproceedings{wei2024unveiling,
  title={Unveiling Selection Biases: Exploring Order and Token Sensitivity in Large Language Models},
  author={Wei, Sheng-Lun and Wu, Cheng-Kuang and Huang, Hen-Hsen and Chen, Hsin-Hsi},
  booktitle={Findings of the Association for Computational Linguistics ACL 2024},
  pages={5598--5621},
  year={2024}
}

@inproceedings{gudatiny,
  title={TINY: Rethinking Selection Bias in LLMs: Quantification and Mitigation using Efficient Majority Voting},
  author={Guda, Blessed and Francis, Lawrence and Ashungafac, Gabrial Zencha and Joe-Wong, Carlee and Busogi, Moise},
  booktitle={ICLR Workshop: Quantify Uncertainty and Hallucination in Foundation Models: The Next Frontier in Reliable AI},
  year = {2025}
}

@inproceedings{chen2024quantifying,
  title={Quantifying and Mitigating Unimodal Biases in Multimodal Large Language Models: A Causal Perspective},
  author={Chen, Meiqi and Cao, Yixin and Zhang, Yan and Lu, Chaochao},
  booktitle={Findings of the Association for Computational Linguistics: EMNLP 2024},
  pages={16449--16469},
  year={2024}
}

@inproceedings{cheng2023can,
  title={Can We Edit Multimodal Large Language Models?},
  author={Cheng, Siyuan and Tian, Bozhong and Liu, Qingbin and Chen, Xi and Wang, Yongheng and Chen, Huajun and Zhang, Ningyu},
  booktitle={Proceedings of the 2023 Conference on Empirical Methods in Natural Language Processing},
  pages={13877--13888},
  year={2023}
}

@inproceedings{wang2024can,
  title={Can We Debias Multimodal Large Language Models via Model Editing?},
  author={Wang, Zecheng and Li, Xinye and Qin, Zhanyue and Li, Chunshan and Tu, Zhiying and Chu, Dianhui and Sui, Dianbo},
  booktitle={Proceedings of the 32nd ACM International Conference on Multimedia},
  pages={3219--3228},
  year={2024}
}

@inproceedings{wang2021gender,
  title={Are Gender-Neutral Queries Really Gender-Neutral? Mitigating Gender Bias in Image Search},
  author={Wang, Jialu and Liu, Yang and Wang, Xin},
  booktitle={Proceedings of the 2021 Conference on Empirical Methods in Natural Language Processing},
  pages={1995--2008},
  year={2021}
}

@inproceedings{gokhale2020mutant,
  title={MUTANT: A Training Paradigm for Out-of-Distribution Generalization in Visual Question Answering},
  author={Gokhale, Tejas and Banerjee, Pratyay and Baral, Chitta and Yang, Yezhou},
  booktitle={Proceedings of the 2020 Conference on Empirical Methods in Natural Language Processing (EMNLP)},
  pages={878--892},
  year={2020}
}

@inproceedings{zhang2024debiasing,
  title={Debiasing multimodal large language models via penalization of language priors},
  author={Zhang, YiFan and Shi, Yang and Yu, Weichen and Wen, Qingsong and Wang, Xue and Yang, Wenjing and Zhang, Zhang and Wang, Liang and Jin, Rong},
  booktitle={Proceedings of the 33rd ACM International Conference on Multimedia},
  pages={4232--4241},
  year={2025}
}

@inproceedings{zong2024fool,
  title={Fool Your (Vision and) Language Model with Embarrassingly Simple Permutations},
  author={Zong, Yongshuo and Yu, Tingyang and Chavhan, Ruchika and Zhao, Bingchen and Hospedales, Timothy},
  booktitle={International Conference on Machine Learning},
  pages={62892--62913},
  year={2024},
  organization={PMLR}
}

@article{cub,
  title={The caltech-ucsd birds-200-2011 dataset},
  author={Wah, Catherine and Branson, Steve and Welinder, Peter and Perona, Pietro and Belongie, Serge},
  journal = {California Institute of Technology},
  year={2011},
  publisher={California Institute of Technology}
}

@inproceedings{dogs,
  title={Novel dataset for fine-grained image categorization: Stanford dogs},
  author={Khosla, Aditya and Jayadevaprakash, Nityananda and Yao, Bangpeng and Li, Fei-Fei},
  booktitle={Proc. CVPR workshop on fine-grained visual categorization (FGVC)},
  volume={2},
  year={2011}
}

@article{aircrafts,
  title={Fine-grained visual classification of aircraft},
  author={Maji, Subhransu and Rahtu, Esa and Kannala, Juho and Blaschko, Matthew and Vedaldi, Andrea},
  journal={arXiv preprint arXiv:1306.5151},
  year={2013}
}

@inproceedings{cars,
  title={3d object representations for fine-grained categorization},
  author={Krause, Jonathan and Stark, Michael and Deng, Jia and Fei-Fei, Li},
  booktitle={Proceedings of the IEEE international conference on computer vision workshops},
  pages={554--561},
  year={2013}
}

@inproceedings{food,
author="Bossard, Lukas
and Guillaumin, Matthieu
and Van Gool, Luc",
editor="Fleet, David
and Pajdla, Tomas
and Schiele, Bernt
and Tuytelaars, Tinne",
title="Food-101 -- Mining Discriminative Components with Random Forests",
booktitle="Computer Vision -- ECCV 2014",
year="2014",
publisher="Springer International Publishing",
  pages={446--461}
}

@inproceedings{finer,
  title={Finer: Investigating and enhancing fine-grained visual concept recognition in large vision language models},
  author={Kim, Jeonghwan and Ji, Heng},
  booktitle={Proceedings of the 2024 Conference on Empirical Methods in Natural Language Processing},
  pages={6187--6207},
  year={2024}
}

@inproceedings{adila2024discovering,
  title = 	 {Discovering Bias in Latent Space: An Unsupervised Debiasing Approach},
  author =       {Adila, Dyah and Zhang, Shuai and Han, Boran and Wang, Bernie},
  booktitle={Proceedings of the 41st International Conference on Machine Learning},
  pages={246--261},
  year={2024},
    volume = 	 {235},
  publisher =    {PMLR}
  
}

@inproceedings{yang2025option,
  title={Option Symbol Matters: Investigating and Mitigating Multiple-Choice Option Symbol Bias of Large Language Models},
  author={Yang, Zhen and Jian, Ping and Li, Chengzhi},
  booktitle={Proceedings of the 2025 Conference of the Nations of the Americas Chapter of the Association for Computational Linguistics: Human Language Technologies (Volume 1: Long Papers)},
  pages={1902--1917},
  year={2025}
}

@misc{qwen3,
      title={Qwen3-VL Technical Report}, 
      author={Shuai Bai and Yuxuan Cai and Ruizhe Chen and Keqin Chen and Xionghui Chen and Zesen Cheng and Lianghao Deng and Wei Ding and Chang Gao and Chunjiang Ge and Wenbin Ge and Zhifang Guo and Qidong Huang and Jie Huang and Fei Huang and Binyuan Hui and Shutong Jiang and Zhaohai Li and Mingsheng Li and Mei Li and Kaixin Li and Zicheng Lin and Junyang Lin and Xuejing Liu and Jiawei Liu and Chenglong Liu and Yang Liu and Dayiheng Liu and Shixuan Liu and Dunjie Lu and Ruilin Luo and Chenxu Lv and Rui Men and Lingchen Meng and Xuancheng Ren and Xingzhang Ren and Sibo Song and Yuchong Sun and Jun Tang and Jianhong Tu and Jianqiang Wan and Peng Wang and Pengfei Wang and Qiuyue Wang and Yuxuan Wang and Tianbao Xie and Yiheng Xu and Haiyang Xu and Jin Xu and Zhibo Yang and Mingkun Yang and Jianxin Yang and An Yang and Bowen Yu and Fei Zhang and Hang Zhang and Xi Zhang and Bo Zheng and Humen Zhong and Jingren Zhou and Fan Zhou and Jing Zhou and Yuanzhi Zhu and Ke Zhu},
      year={2025},
      eprint={2511.21631},
      archivePrefix={arXiv},
      primaryClass={cs.CV},
      url={https://arxiv.org/abs/2511.21631}, 
}

@inproceedings{ma2023poe,
  title={POE: Process of Elimination for Multiple Choice Reasoning},
  author={Ma, Chenkai and Du, Xinya},
  booktitle={Proceedings of the 2023 Conference on Empirical Methods in Natural Language Processing},
  pages={4487--4496},
  year={2023}
}

@inproceedings{fu2025exclusion,
  title={Exclusion of Thought: Mitigating Cognitive Load in Large Language Models for Enhanced Reasoning in Multiple-Choice Tasks},
  author={Fu, Qihang and Qin, Yongbin and Huang, Ruizhang and Chen, Yanping and Zhou, Yulin and Long, Lintao},
  booktitle={Proceedings of the 63rd Annual Meeting of the Association for Computational Linguistics (Volume 1: Long Papers)},
  pages={21673--21686},
  year={2025}
}

@inproceedings{zhao2025largelanguagemodelsbadly,
  title={Large language models badly generalize across option length, problem types, and irrelevant noun replacements},
  author={Zhao, Guangxiang and Hu, Saier and Jian, Xiaoqi and Jinzhu, Wu and Wu, Yuhan and Sun, Lin and Zhang, Xiangzheng},
  booktitle={Proceedings of the 2025 Conference on Empirical Methods in Natural Language Processing},
  pages={26825--26834},
  year={2025}
}

@inproceedings{atabuzzaman2025,
  title={Zero-shot Fine-Grained Image Classification using Large Vision-Language Models},
  author={Atabuzzaman, Md and Zhang, Andrew and Thomas, Chris},
  booktitle={Findings of the Association for Computational Linguistics: EMNLP 2025},
  pages = "23569--23582",
  year={2025}
}

@article{atabuzzaman2025benchmarking,
  title={Benchmarking and Mitigating {MCQA} Selection Bias of Large Vision-Language Models},
  author={Atabuzzaman, Md and Asgarov, Ali and Thomas, Chris},
  journal={Proceedings of the 2025 Conference on Empirical Methods in Natural Language Processing},
  pages = "33548--33562",
  year={2025}
}

@inproceedings{holtzman2021surface,
  title={Surface form competition: Why the highest probability answer isn’t always right},
  author={Holtzman, Ari and West, Peter and Shwartz, Vered and Choi, Yejin and Zettlemoyer, Luke},
  booktitle={Proceedings of the 2021 conference on empirical methods in natural language processing},
  pages={7038--7051},
  year={2021}
}

\appendix

\section{Appendix}
\label{sec:appendix}

%------------------------------------------
This section contains the details of the following topics.
\begin{itemize}[noitemsep, topsep=0pt]
    \item Dataset and Dataset Generation Constraints (Appendix \ref{dataset_constraints})
    \item Effect of Visual Degradation on Conflict Resolution (Appendix \ref{sec:effect_blur})
    \item Architectural Localization via Attention Analysis (Appendix \ref{sec:attention_analysis})
    \item Caption-Guided MCQA Prompt (Appendix \ref{sec:caption_mcqa})
    \item Caption-Guided MCQA (Appendix \ref{sec:caption_guided_mcqa})
    \item Yes/No Question Answering  (Appendix \ref{sec:yes_no_qa})
    \item MCQA with Selection by Rejection (Appendix \ref{sec:selection_by_rejection})
    \item LVLMs Performance Under Different Option-Length Conditions (Appendix \ref{lvlm_performance})
    \item Results for the Abstention (NoA) Settings (Appendix \ref{NoA_results})
    \item Position Bias and Answer Distribution Analysis (Appendix \ref{sec:position_bias})
\end{itemize}

%%%%%%%%%%%%%%%%%%%%%%%%%%%%%%%%%%%%%%%%%%%%%%%%%%%%

\subsection{Dataset and Dataset Generation Constraints}
\label{dataset_constraints}

\begin{figure}[!ht]
    \centering
    \includegraphics[width=\linewidth]{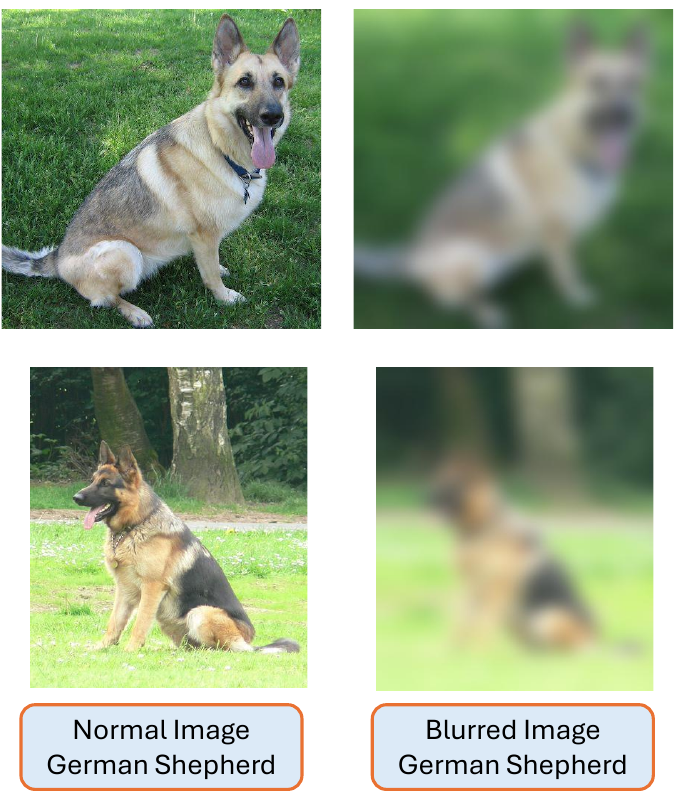}
    \caption{Example of normal and blurred image of German Shepherd.}
    \label{fig:ex_image}
\end{figure}

We enforced the following generation constraints to ensure semantic and visual fidelity of the length-controlled variants.

\textbf{Semantic preservation}: We rephrase options only to adjust length; the factual meaning and fine-grained class identity remain unchanged.
    
\textbf{Visual consistency}: We ground all modifications in visual attributes present in the image. We introduce no information absent from the base question or image.

\textbf{Structural consistency}: We ensure all options within a variant share the same fixed prefix (e.g., "This image shows..." or "This is..."), with length variations applied only to the descriptive content following the prefix.

\textbf{Natural phrasing}: We use natural, grammatically correct language to avoid awkward constructions that might signal correctness independent of content.
    
\textbf{Class structure preservation}: Each option continues to represent a distinct fine-grained class, maintaining the original task difficulty.

\paragraph{Verification of Dataset Fidelity.}

To verify that GPT-4o rewrites preserved semantic correctness and visual fidelity, we conducted structured manual inspection of stratified samples from every dataset. For each rewritten option, we verified that: (1) all class-defining visual attributes were preserved, (2) no contradictory or hallucinated attributes were introduced, and (3) the class identity remained unambiguous relative to distractors. All examined samples satisfied these criteria.

We further validated fidelity through two indirect measures. First, the \textbf{Equal Long} and \textbf{Equal Short} variants yield broadly comparable accuracies for most models and domains (see Table~\ref{tab:model_length_bias_comparison}), with a mean absolute difference of approximately 5 points; notable exceptions include LLaVA-v1.5 variants on Food, where gaps reach up to 28 points, likely reflecting model-specific sensitivity to phrasing length rather than dataset construction artifacts. Critically, even these largest baseline gaps remain substantially smaller than the length bias effect itself: for example, GPT-4o's Equal Long--Equal Short gap on CUB is 8.34 points, compared to a 47.66-point gap between Shorter Correct and Longer Correct (Table~\ref{tab:gpt_model}). This indicates that residual variation from the rewriting process is minor relative to the systematic bias our benchmark is designed to measure. Second, if GPT-4o had introduced systematic bias during rewriting, GPT-4o's evaluation performance would differ noticeably from models uninvolved in dataset creation. Instead, GPT-4o exhibits length bias patterns closely matching o4-mini and Claude Sonnet-4.5 (Table~\ref{tab:gpt_model}), confirming that the rewritten dataset is semantically faithful and unbiased.

\begin{figure*}[!ht]
    \centering
    \includegraphics[width=\linewidth]{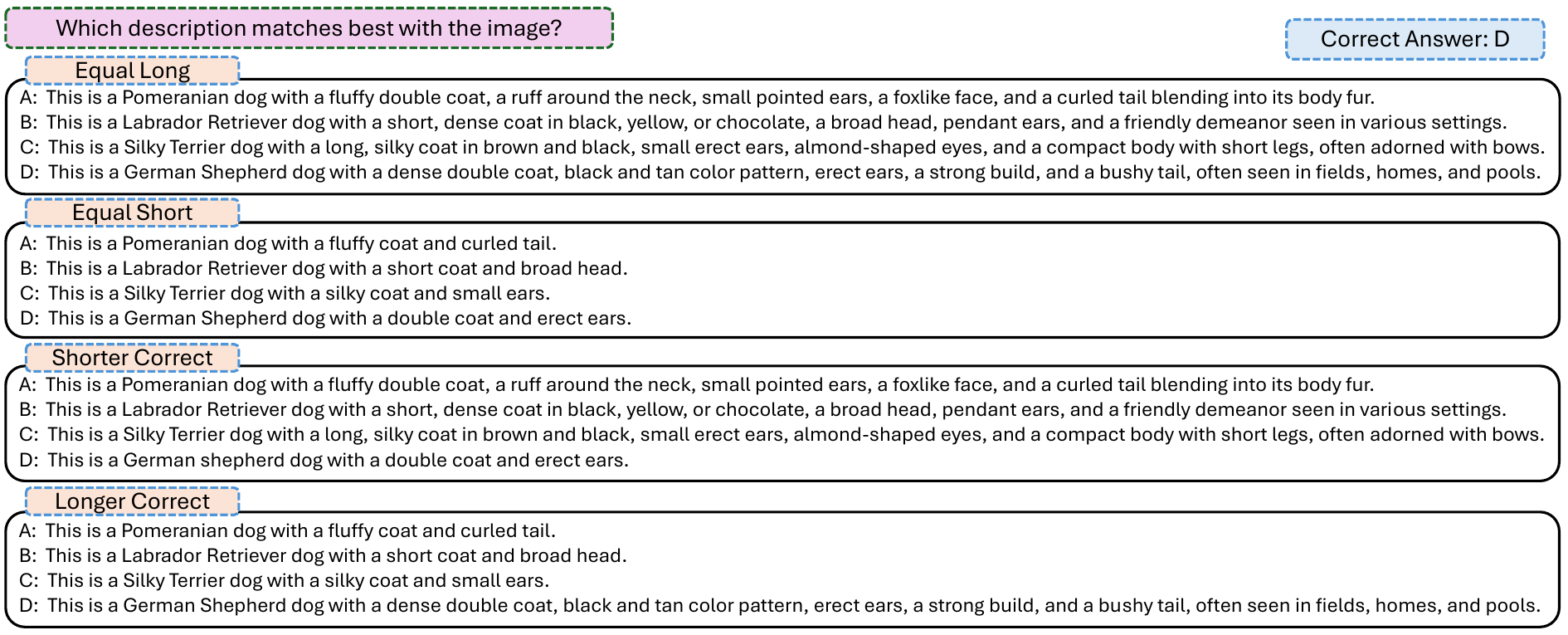}
    \caption{Example multiple-choice questions from the Dogs dataset illustrating four length-controlled variants: \textbf{Equal Long}, \textbf{Equal Short}, \textbf{Shorter Correct}, and \textbf{Longer Correct}. The correct answer (D: German Shepherd) remains constant while option lengths vary systematically to isolate the effect of textual length on model predictions.}
    \label{fig:four_variants}
\end{figure*}

Figure~\ref{fig:four_variants} illustrates an example from our length-controlled MCQA benchmark. Each variant manipulates the relative length of the correct answer and distractors while preserving semantic content and visual accuracy. In \textbf{Equal Long} and \textbf{Equal Short}, all options have approximately the same length, serving as balanced baselines. In \textbf{Shorter Correct}, the correct answer (D) is concise while distractors (A, B, C) are verbose, testing whether models exhibit bias against selecting shorter options. Conversely, in \textbf{Longer Correct}, the correct answer is detailed while distractors are brief, testing preference for longer options. These controlled manipulations enable direct measurement of how answer length influences LVLM selection independent of visual or semantic factors.

Figure \ref{fig:ex_image} illustrates two German Shepherd examples from the Dogs dataset: one original and one blurred version used for robustness evaluation.

%%%%%%%%%%%%%%%%%%%%%%%%%%%%%%%%%%%%%%%%%%%%%%%%%%%%%%%%

%%%%%%%%%%%%%%%%%%%%%%%%%%%%%%%%%%%
\begin{figure*}[!ht]
    \centering
    \includegraphics[width=\linewidth]{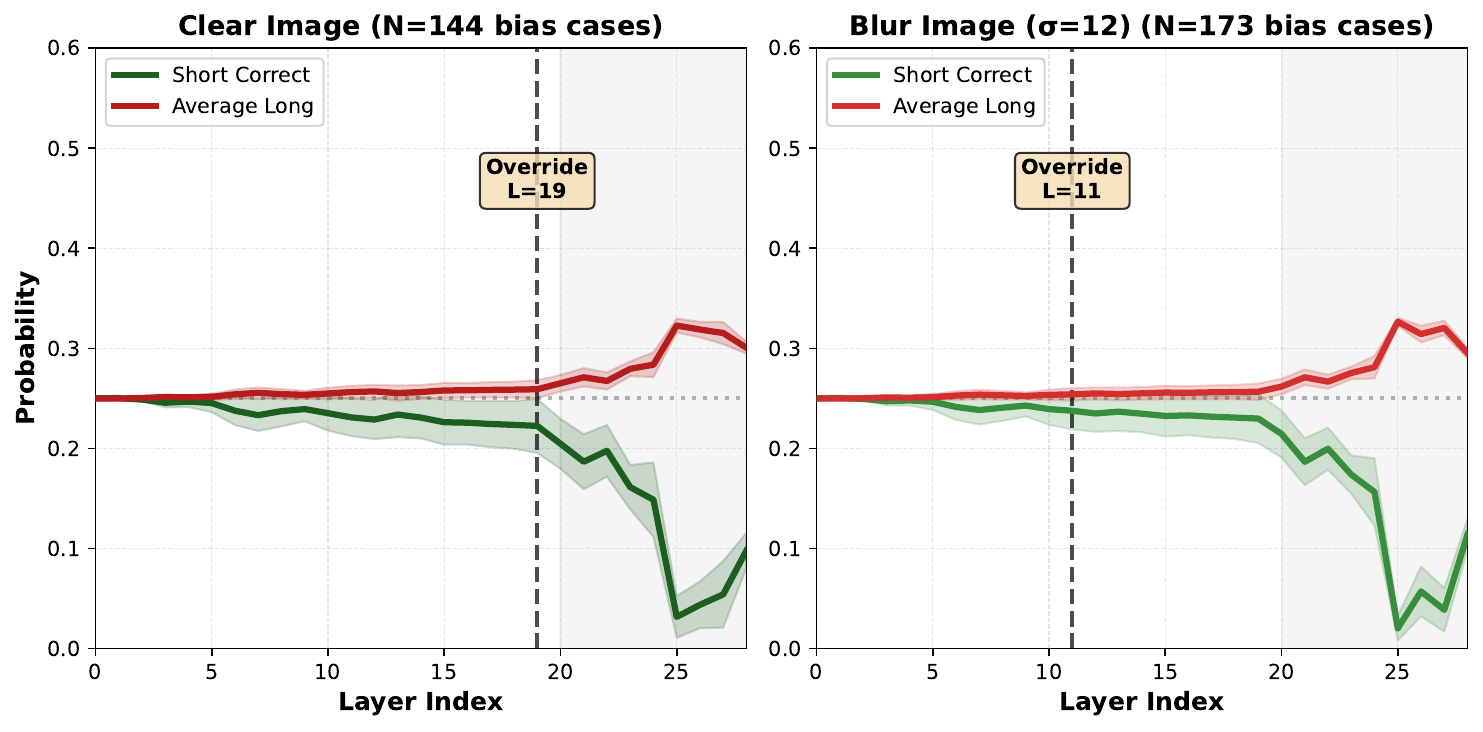}
    \vspace{-0.75cm}
    \caption{\textbf{Effect of visual degradation on length-bias conflict resolution.} Layer-wise probabilities for the short correct answer (green) and the average long distractor (red) under clear images (left) and Gaussian blur ($\sigma = 12$, right), evaluated on the same set of length-biased error cases. Dashed lines mark the median override layer under each condition. Clear images exhibit greater dispersion in probability trajectories due to heterogeneity in visual evidence strength across examples, while blurring homogenizes visual inputs and produces earlier, more uniform dominance of length bias. Shaded regions denote 95\% confidence intervals.}
    \label{fig:blur}
\end{figure*}

%%%%%%%%%%%%%%%%%%%%%%%%%%%%%%%%%%%%%%%%%%%%%%%%%%%%%%%%%%

\subsection{Effect of Visual Degradation on Conflict Resolution}
\label{sec:effect_blur}
To causally probe the role of visual evidence in length-biased decision making, we repeat our layer-wise analysis under controlled visual degradation. Specifically, we apply Gaussian blur ($\sigma = 12$) to the input images while keeping the question, answer options, and decoding procedure unchanged. This manipulation weakens fine-grained visual cues without altering linguistic content, allowing us to isolate the effect of visual signal strength on internal conflict resolution dynamics. We define the \emph{override layer} as the first layer where the average long distractor probability exceeds the short correct answer probability for three consecutive layers, indicating stable dominance of length bias over visual evidence. This statistic is distinct from the \emph{stable inversion layer} ($L=14$) reported in Section~\ref{sec:layerwise_analysis}: the override layer is computed only on length-biased error cases using the averaged probability across all three long distractors, whereas the stable inversion layer is computed across the full Shorter Correct split using a single fixed (longest) distractor.

Figure~\ref{fig:blur} compares layer-wise probability trajectories for clear versus blurred images on the same set of length-biased error cases. Under clear images, visual evidence delays the dominance of long distractors until late layers (median override layer $L = 19$). When visual inputs are blurred, this dominance emerges substantially earlier (median override layer $L = 11$), indicating that weakened visual evidence accelerates the resolution of conflict in favor of linguistic priors.

Notably, probability trajectories under clear images exhibit greater dispersion than those under blur. This reflects heterogeneity in visual evidence strength across examples: clear images preserve variation in pose, viewpoint, and fine-grained discriminability, allowing some instances to sustain visual alignment longer than others. Blurring homogenizes visual inputs across examples, reducing this variability and resulting in more uniform, earlier dominance of length bias. The systematic shift in override timing and reduced dispersion under blur provide causal evidence that late-layer conflict resolution in LVLMs depends on the strength of visual evidence, rather than arising solely from inherited language-model biases.

Together, these results demonstrate that visual information can delay—but does not reliably prevent—the amplification of length bias. When visual evidence is degraded, linguistic priors dominate earlier in the computation, confirming that length bias in LVLMs emerges from dynamic competition between modalities rather than from static decoder preferences alone.

%%%%%%%%%%%%%%%%%%%%%%%%%%%%%%%%%%%%%%%%%%%%%%%%%%%%%%%%%%%

\subsection{Architectural Localization via Attention Analysis}
\label{sec:attention_analysis}

\begin{figure}[!ht]
    \centering
    \includegraphics[width=\linewidth]{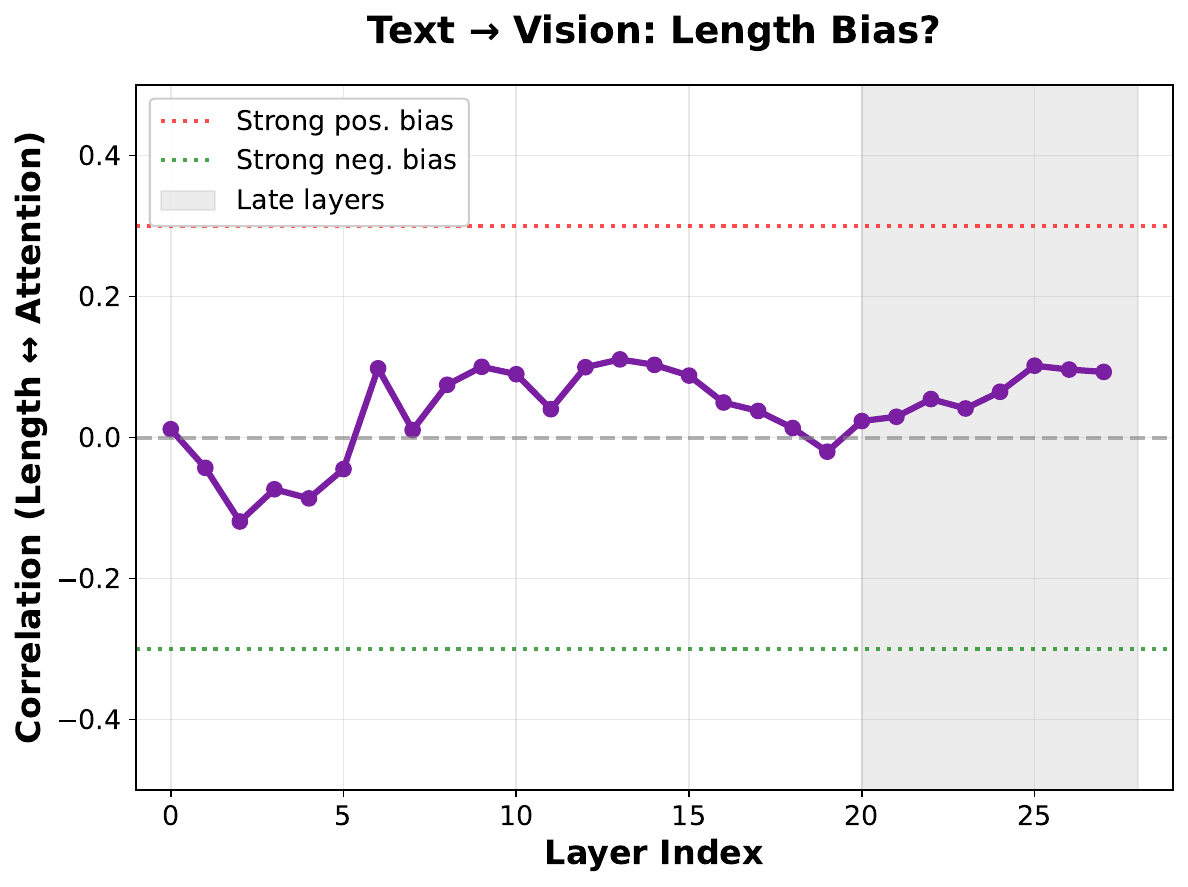}
    \caption{Text-to-vision attention correlation across transformer layers (N=200 examples). We measure the correlation between option length and visual attention at each layer. Late layers (20--28, shaded region) show statistically significant positive correlation (mean $r = 0.063$, $p < 0.001$), indicating that longer options query vision more than shorter options. Despite this increased visual grounding, long incorrect options are preferentially selected, demonstrating that length bias operates through override of visual evidence in the language decoder rather than insufficient visual information retrieval.}
    \label{fig:attention}
    \vspace{-0.5cm}
\end{figure}

To identify whether length bias originates in vision-language fusion or language processing, we analyze text-to-vision attention patterns across all layers (N=200 examples). We extract attention weights from text option tokens to vision tokens, measuring whether longer options differentially query visual information during decision-making.

Late-layer analysis (layers 20--28, Figure~\ref{fig:attention}) reveals a statistically significant positive correlation between option length and visual attention ($r = 0.063$, $p < 0.001$). Critically, long incorrect options attend 9\% more to visual features than short correct answers (0.000342 vs 0.000313, $p < 0.001$), yet are still preferentially selected. This demonstrates that length bias does not arise from inadequate visual grounding but rather from the language decoder systematically overriding visual evidence when processing longer text sequences.

This localization explains why caption-guided MCQA (Section~\ref{sec:caption_guided_mcqa}) provides only partial mitigation: explicit visual descriptions cannot overcome bias that operates independently of visual evidence strength. The finding indicates that effective solutions must target the language decoder through architectural modifications or length-normalized scoring mechanisms.

%%%%%%%%%%%%%%%%%%%%%%%%%%%%%%%%%%%%%%%%%%%%%%%%%%%%%%%%%%%

\subsection{Caption-Guided MCQA Prompt}
\label{sec:caption_mcqa}
For caption-guided MCQA experiments, we use the following prompt template (\ref{fig:caption_mcqa_prompt}) to guide the answering model to ground its reasoning in both the image and the generated caption:

\begin{figure}[!ht]
\centering
\begin{tcolorbox}[
  enhanced,
  colback=gray!5,
  colframe=gray!75!black,
  title=Caption-Guided MCQA Prompt,
  fonttitle=\bfseries,
  listing only,
  listing engine=listings,
  listing options={
    basicstyle=\ttfamily\small,
    breaklines=true,
    breakatwhitespace=false,
    columns=fullflexible,
    keepspaces=true,
    postbreak=\mbox{\textcolor{gray}{$\hookrightarrow$}\space}
  }
]
You are given a multiple-choice question about an image.\\
Image Description: \{image\_description\}\\
Use BOTH the image and the description above to answer the 
question. Return ONLY a single option ID from ('A', 'B', 'C', 
'D') of the correct answer. Do not include explanations or 
extra text.
\{formatted\_prompt\}\\
Answer: (choose one from A, B, C, D)
\end{tcolorbox}
\vspace{-0.5cm}
\caption{Prompt template used for caption-guided MCQA.}
\label{fig:caption_mcqa_prompt}
\end{figure}

\noindent where \texttt{\{image\_description\}} is replaced with the generated caption from the captioning model, and \texttt{\{formatted\_prompt\}} contains the question and answer options.

%%%%%%%%%%%%%%%%%%%%%%%%%%%%%%%%%%%%%%%%%%%%%%%%%%%%%%%%%%%%%%%%%%%%%%%%%

\begin{table}[!t]
\centering
\begingroup
\setlength{\tabcolsep}{3.2pt} % tighter spacing
\renewcommand{\arraystretch}{0.95}
\scriptsize
\resizebox{\columnwidth}{!}{%
\begin{tabular}{lllcccc}
\toprule
\textbf{Dataset} & \textbf{Captioner} & \textbf{Model} &
\textbf{\makecell{Equal\\Long}} & \textbf{\makecell{Equal\\Short}} &
\textbf{\makecell{Shorter\\Correct}} & \textbf{\makecell{Longer\\Correct}} \\
\midrule

\multirow{6}{*}{\textbf{CUB}} 
  & Qwen2.5-VL-7B   & Qwen2.5-VL-7B   & 59.30 & 52.10 & 41.30 & 70.50 \\
  & Qwen3-VL-30B    & Qwen2.5-VL-7B   & 66.25	& 55.50	& 44.25	& 70.00 \\
  & GPT-4o          & Qwen2.5-VL-7B   & \textbf{67.00} & \textbf{58.00} & \textbf{48.50} & \textbf{70.75} \\ \cmidrule(lr){2-7}
  & InternVL-2.5-8B & InternVL-2.5-8B & 64.50 & 55.10 & 41.80 & 74.40 \\
  & Qwen3-VL-30B    & InternVL-2.5-8B & \textbf{66.75}	& \textbf{57.50}	& \textbf{46.00}	& \textbf{78.25} \\
  & GPT-4o          & InternVL-2.5-8B & 65.25 & 57.25 & 43.50 & 76.75 \\ 
\midrule

\multirow{6}{*}{\textbf{Dogs}} 
  & Qwen2.5-VL-7B   & Qwen2.5-VL-7B   & 57.33 & 51.50 & 53.00 & \textbf{61.17} \\
  & Qwen3-VL-30B    & Qwen2.5-VL-7B   & 65.00	& \textbf{56.67}	& \textbf{58.33}	& 60.83 \\
  & GPT-4o          & Qwen2.5-VL-7B   & \textbf{67.08} & 53.33 & 56.67 & 59.58 \\ \cmidrule(lr){2-7}
  & InternVL-2.5-8B & InternVL-2.5-8B & 56.33 & 52.33 & 42.33 & 68.83 \\
  & Qwen3-VL-30B    & InternVL-2.5-8B & \textbf{61.67}	& \textbf{59.58}	& \textbf{50.42}	& \textbf{71.67} \\
  & GPT-4o          & InternVL-2.5-8B & 60.83 & 57.50 & 47.50 & 68.75 \\
\bottomrule
\end{tabular}%
}
\endgroup
\vspace{-0.25cm}
\caption{
Accuracy (\%) of LVLMs on \textbf{Caption-Guided MCQA} across CUB and Dogs without class names in the answer options.}
\vspace{-0.5cm}
\label{tab:caption_guided}
\end{table}

\subsection{Caption-Guided MCQA}
\label{sec:caption_guided_mcqa}

To investigate whether bias arises from visual perception or decision-making, we introduce \textbf{Caption-Conditioned MCQA (CC-MCQA)} inspired by human problem-solving: first forming an internal description of the scene, then comparing it against candidate answers. This framework decouples perception from answer selection through two sequential stages, serving as both a mitigation strategy and diagnostic tool.

\noindent\textbf{Caption Generation (Perception Stage).}
Given an image $I$, a captioning model $M_{\text{cap}}$ generates a fine-grained description:
\[
C = M_{\text{cap}}(I, p_{\text{cap}}),
\]
where $p_{\text{cap}}$ encourages grounding of discriminative attributes (e.g., color, pattern, texture) while discouraging speculative language.

\noindent\textbf{Caption-Guided Answer Selection (Reasoning Stage).}
The caption $C$ is provided to an answering model $M_{\text{ans}}$ with the question $Q$ and candidate options $\{O_1, \dots, O_n\}$:
\[
\hat{y} = M_{\text{ans}}(I, Q, \{O_1, \dots, O_n\}, C, p_{\text{ans}}),
\]
where $p_{\text{ans}}$ guides the model to ground reasoning in the caption rather than option length. This disentangles perception errors (from $M_{\text{cap}}$) from decision biases (from $M_{\text{ans}}$).

We evaluate two variants: \textbf{Self-Conditioned} ($M_{\text{cap}} = M_{\text{ans}}$) and \textbf{Decoupled} ($M_{\text{cap}} \ne M_{\text{ans}}$).  
If performance improves with accurate captions, biases originate from perception errors; if performance degrades despite accurate captions, biases stem from decision-making heuristics.

%%%%----------------------------------------

\paragraph{Results.}
Table~\ref{tab:caption_guided} presents Caption-Guided MCQA performance without class names. We compare against baseline (Table~\ref{tab:model_length_bias_comparison}) to assess whether explicit visual grounding mitigates length bias.

\noindent\textbf{Caption Guidance Substantially Reduces Length Bias.}
Caption-guided MCQA with GPT-4o substantially narrows the gap between "Shorter Correct" and "Longer Correct." For Qwen2.5-VL-7B on CUB, the baseline gap of 66.80 points (26.60\% vs 93.40\%) reduces to 22.25 points (48.50\% vs 70.75\%), a 66.7\% bias reduction. The decrease in "Longer Correct" (93.40\% to 70.75\%) represents correction of artificially inflated accuracy, while "Shorter Correct" improves substantially (26.60\% to 48.50\%), demonstrating that visual grounding overcomes bias against concise options.

\noindent\textbf{External Captions Outperform Self-Conditioned Approaches.}
Using stronger captioners (GPT-4o, Qwen3-VL-30B) consistently outperforms both self-conditioned and baseline approaches. On CUB "Equal Long," GPT-4o achieves 67.00\% versus 59.30\% (self-conditioned) and 64.90\% (baseline). Self-conditioned captioning often degrades performance, revealing that perception errors compound decision-making biases when the same model handles both stages.

\noindent\textbf{Residual Bias Reveals Multi-Stage Architecture.}
Despite substantial improvements, a 22.25 point gap persists even with GPT-4o captions, revealing that length preference operates at multiple processing stages. While explicit visual grounding addresses perceptual biases, decision-making biases in textual weighting remain partially intact.

%%%%%%%%%%%%%%%%%%%%%%%%%%%%%%%%%%%%%%%%%%%%%%%%%%%%%%%%

%%%%%%%%%%%%%%%%%%%%%%%%%%%%%%%%%%%%%%%%%%%%%%%%%%%%%%%%%%%%%%%%%%%%%%
\begin{table}[!t]
\centering
\begingroup
\setlength{\tabcolsep}{3.2pt} % tighten column spacing
\renewcommand{\arraystretch}{0.95}
\scriptsize
\resizebox{\columnwidth}{!}{%
\begin{tabular}{llccccc}
\toprule
\textbf{Dataset} & \textbf{Model} &
\textbf{\makecell{Equal\\Long}} & \textbf{\makecell{Equal\\Short}} &
\textbf{\makecell{Shorter\\Correct}} & \textbf{\makecell{Longer\\Correct}} \\
\midrule

\multirow{2}{*}{CUB (W/o)} 
  & Qwen2.5-VL-7B & 58.40 & 51.30 & 62.80 & \textbf{45.70} \\
  & InternVL\_2.5-8B & \textbf{59.00} & \textbf{53.50} & \textbf{67.10} & 44.00 \\ 
\cmidrule(lr){1-6}

\multirow{2}{*}{Dogs (W/o)} 
  & Qwen2.5-VL-7B & \textbf{57.17} & \textbf{53.17} & 53.00 & \textbf{61.33} \\
  & InternVL\_2.5-8B & 52.00 & 52.00 & \textbf{73.33} & 32.50 \\ 
\midrule

\multirow{2}{*}{CUB (W.)} 
  & Qwen2.5-VL-7B & \textbf{71.20} & \textbf{68.60} & \textbf{69.10} & \textbf{70.40} \\
  & InternVL\_2.5-8B & 63.60 & 62.00 & 53.20 & 70.10 \\ 
\cmidrule(lr){1-6}

\multirow{2}{*}{Dogs (W.)} 
  & Qwen2.5-VL-7B & \textbf{71.83} & \textbf{71.17} & \textbf{77.33} & \textbf{61.83} \\
  & InternVL\_2.5-8B & 55.17 & 57.33 & 67.17 & 44.00 \\
\bottomrule
\end{tabular}%
}
\endgroup
\vspace{-0.25cm}
\caption{Accuracy (\%) of LVLMs on "\textbf{Yes/No}" evaluation across CUB and Dogs, reported with (W.) and without (W/o) class names in the answer options.}
\label{tab:ryes_no_image}
\end{table}

%%%%------------------------------------------------------

\subsection{Yes/No Question Answering (QA)}
\label{sec:yes_no_qa}
We transform the multiple-choice task into binary Yes/No questions. For each option, we ask the model whether it correctly answers the question and select the option with the highest probability assigned to "Yes" as the predicted answer. This independent evaluation removes direct comparisons between options during selection.

Table~\ref{tab:ryes_no_image} presents results with and without class names. The Yes/No format substantially improves "Shorter Correct" performance: for Qwen2.5-VL-7B on CUB with class names, accuracy reaches 69.10\% compared to 26.60\% baseline (Table~\ref{tab:model_length_bias_comparison}), nearly achieving parity with "Longer Correct" (70.40\%). However, performance degrades in balanced settings, with "Equal Short" dropping from 61.00\% to 51.30\%. While independent option evaluation mitigates length bias by removing direct length comparisons, eliminating comparative context hinders fine-grained discrimination when length is not a confounding factor.

%%%%%%%%%%%%%%%%%%%%%%%%%%%%%%%%%%%%%%%%%%%%%%%%%%%%%%%%%%%%%%%%%%%%%%%%%%
\begin{table}[!t]
\centering
\begingroup
\setlength{\tabcolsep}{3.2pt}
\renewcommand{\arraystretch}{0.95}
\scriptsize
\resizebox{\columnwidth}{!}{%
\begin{tabular}{llccccc}
\toprule
\textbf{Dataset} & \textbf{Model} &
\textbf{\makecell{Equal\\Long}} & \textbf{\makecell{Equal\\Short}} &
\textbf{\makecell{Shorter\\Correct}} & \textbf{\makecell{Longer\\Correct}} \\
\midrule

\multirow{2}{*}{CUB (W/o)} 
  & Qwen2.5-VL-7B & 52.80 & \textbf{52.40} & 31.40 & 81.80 \\
  & InternVL\_2.5-8B & \textbf{56.80} & 49.80 & \textbf{33.10} & \textbf{82.90} \\ 
\cmidrule(lr){1-6}

\multirow{2}{*}{Dogs (W/o)} 
  & Qwen2.5-VL-7B & \textbf{54.17} & 53.83 & \textbf{41.50} & 73.50 \\
  & InternVL\_2.5-8B & 54.00 & \textbf{56.33} & 32.67 & \textbf{76.67} \\ 
\midrule

\multirow{2}{*}{CUB (W.)} 
  & Qwen2.5-VL-7B & \textbf{66.70} & \textbf{63.40} & \textbf{50.00} & 84.50 \\
  & InternVL\_2.5-8B & 57.90 & 56.00 & 33.40 & \textbf{88.10} \\ 
\cmidrule(lr){1-6}

\multirow{2}{*}{Dogs (W.)} 
  & Qwen2.5-VL-7B & \textbf{69.83} & \textbf{70.17} & \textbf{60.17} & \textbf{89.67} \\
  & InternVL\_2.5-8B & 58.67 & 52.83 & 38.17 & 78.00 \\
\bottomrule
\end{tabular}%
}
\endgroup
\vspace{-0.25cm}
\caption{Accuracy (\%) of LVLMs on "\textbf{selection by rejection}" across CUB and Dogs, with (W.) and without (W/o) class names in the answer options.}
\vspace{-0.5cm}
\label{tab:rejection_image}
\end{table}

%%%%------------------------------------------------------

\subsection{MCQA with Selection by Rejection}
\label{sec:selection_by_rejection}

In multiple-choice reasoning, humans often approach difficult questions by first eliminating clearly incorrect options before selecting a final answer. Inspired by this process, we design a \textbf{\textit{selection-by-rejection}} strategy to test whether explicit elimination can mitigate the length bias.

Instead of prompting the model to select a single option directly, we instruct it to \textit{reject} three candidates, inferring the final prediction as the surviving option. We hypothesize that explicitly discarding incorrect candidates might reduce reliance on superficial heuristics such as option length and instead focus on genuine grounding in the image. 

\noindent\textbf{Selection-by-Rejection Degrades Performance and Reinforces Length Bias.}
Table~\ref{tab:rejection_image} shows that this strategy consistently degrades performance across all models. Comparing with Table~\ref{tab:model_length_bias_comparison}, we observe substantial accuracy drops. On CUB without class names, Qwen2.5-VL-7B drops from 64.90\% to 52.80\% in "Equal Long" and from 93.40\% to 81.80\% in "Longer Correct." Although LVLMs produce syntactically valid rejections, the eliminations are not reliably aligned with correctness. Further analysis reveals that generated justifications are typically plausible-sounding but not systematically grounded in visual or textual evidence. Unlike humans, who perform elimination as a structured logical process, LVLMs treat rejection as another generative task influenced by the same shallow heuristics. This suggests that the rejection procedure does not reduce length bias but instead reinforces the preference for visually inconsistent longer options.

%%%%%%%%%%%%%%%%%%%%%%%%%%%%%%%%%%%%%%%%%%%%%%%%%%%%%%%%%%%%%%%%%%%%%%%%%%%%%%
\subsection{LVLM Performance Under Different Option-Length Conditions}
\label{lvlm_performance}

Table~\ref{tab:model_length_bias_comparison} presents comprehensive results across all evaluated models and datasets under four length-controlled conditions. The $\Delta$ column quantifies length bias magnitude, showing substantial bias across all models: without class names, $\Delta$ ranges from 8.52 (InternVL\_2.5-8B on Food) to 93.50 (LLaVA-v1.5-7B on Dogs). Performance consistently drops in "Shorter Correct" conditions (negative values in parentheses) and increases in "Longer Correct" conditions (positive values), with drops of 2–38 points and gains of 1–70 points depending on model and dataset. Class name inclusion mitigates bias, reducing $\Delta$ values substantially: for Qwen2.5-VL-7B on CUB, $\Delta$ decreases from 66.80 to 43.20. However, even with class names, significant bias persists (e.g., LLaVA-v1.5-13B on CUB: $\Delta$ = 57.30), indicating that length preference is deeply entrenched. Notably, the Food dataset shows lower bias across all models, suggesting that domain characteristics influence susceptibility to length-based heuristics.

\begin{table*}[!ht]
\centering
\small
\begin{tabular}{lllccccc}
\toprule
\textbf{\makecell{Class\\Name}} &\textbf{Dataset} & \textbf{Model} & \textbf{\makecell{Equal\\Long}} & \textbf{\makecell{Equal\\Short}} & \textbf{\makecell{Shorter\\Correct}} & \textbf{\makecell{Longer\\Correct}} & $\Delta$ \\
\midrule

 \multirow{28}{*}{Without} &\multirow{4}{*}{Aircraft} & LLaVA-v1.5-7B & 32.86 & 34.00 (+1.14) & 18.29 (-14.57) & 84.57 (+51.71) &66.28 \\
 & & LLaVA-v1.5-13B & 39.71 & 35.71 (-4.00) & 18.86 (-20.85) & \textbf{87.43} (+47.72) & 68.57 \\
 & & Qwen2.5-VL-7B & \textbf{54.86} & \textbf{43.71} (-11.15) & 21.43 (-33.43) & 84.00 (+29.14) & 62.57\\
 & & InternVL\_2.5-8B & 39.71 & 36.29 (-3.42) & \textbf{32.29} (-7.42) & 57.43 (+17.72) & \textbf{25.14} \\ \cmidrule(lr){2-8}

 &\multirow{6}{*}{Cars} & LLaVA-v1.5-7B & 26.73  & 32.65 (+5.92)  & 12.35 (-14.38) & \textbf{94.29} (+67.56) & 81.94 \\
 & & LLaVA-v1.5-13B & 40.71 & 44.08 (+3.37) & 16.94 (-23.77) & 93.16 (+52.45) & 76.22 \\
 & & Qwen2.5-VL-3B & 64.80 & 55.51 (-9.29) & 35.82 (-28.98) & 85.61 (+20.81) & 49.79 \\
 & & Qwen2.5-VL-7B & 68.88 & 59.59 (-9.29) & 38.57 (-30.31) & 89.69 (+20.81) & 51.12 \\
 & & Qwen2.5-VL-32B & \textbf{73.16} & \textbf{60.71} (-12.45) & \textbf{42.86} (-30.30) & 92.65 (+19.49) & 49.79 \\
 & & InternVL\_2.5-8B & 56.94 & 47.45 (-9.49) & 37.55 (-19.39) & 72.65 (+15.71) & \textbf{35.10} \\ \cmidrule(lr){2-8}
 
 &\multirow{7}{*}{CUB} & LLaVA-v1.5-7B & 26.70 & 31.20 (+4.50) & 8.90 (-17.80) & \textbf{96.80} (+70.10) & 87.90 \\
 & & LLaVA-v1.5-13B & 39.30 & 42.60 (+3.30) & 15.20 (-24.10) & 93.30 (+54.00) & 78.10 \\
 & & Qwen2.5-VL-3B & 60.60 & 58.40 (-2.20) & 38.10 (-22.50) & 81.40 (+20.80) & 43.30 \\
 & & Qwen2.5-VL-7B & 64.90 & 61.00 (-3.90) & 26.60 (-38.30) & 93.40 (+28.50) & 66.80 \\
 & & Qwen2.5-VL-32B & 68.00 & 61.30 (-6.70) & 37.20 (-30.80) & 93.30 (+25.30) & 56.10 \\
 & & Qwen3-VL-30B   & \textbf{72.30}	& \textbf{63.20} (-9.10) & \textbf{47.30} (-25.00)	& 85.00 (+12.70) & \textbf{37.70} \\
 & & InternVL\_2.5-8B & 62.50 & 57.10 (-5.40) & 37.80 (-24.70) & 88.30 (+25.80) & 50.50 \\ \cmidrule(lr){2-8}

 &\multirow{5}{*}{Dogs} & LLaVA-v1.5-7B & 30.00 & 33.67 (+3.67)  & 5.67 (-24.33)  & \textbf{99.17} (+69.17) & 93.50 \\
 & & LLaVA-v1.5-13B & 36.67 & 42.33 (+5.66)  & 14.00 (-22.67) & 95.67 (+59.00) & 81.67 \\
 & & Qwen2.5-VL-7B & 62.83 & 58.50 (-4.33)  & 33.17 (-29.66) & 90.83 (+28.00) & 57.66 \\
 & & Qwen3-VL-30B   & \textbf{71.17}	& \textbf{64.00} (-7.17)	 & \textbf{49.50} (-21.67)	& 89.17 (+18.00) & \textbf{39.67} \\
 & & InternVL\_2.5-8B & 57.50 & 58.00 (+0.50)  & 38.17 (-19.33) & 79.67 (+22.17) & 41.50 \\ \cmidrule(lr){2-8}

 &\multirow{4}{*}{Food} & LLaVA-v1.5-7B & 33.47 & 53.07 (+19.60) & 27.72 (-5.75)  & 93.07 (+59.60) & 65.35 \\
 & & LLaVA-v1.5-13B & 49.50 & 65.15 (+15.65) & 42.57 (-6.93)  & \textbf{95.84} (+46.34) & 53.27 \\
 & & Qwen2.5-VL-7B & \textbf{86.73} & \textbf{86.93} (+0.20)  & \textbf{78.61} (-8.12)  & 92.08 (+5.35) & 13.47 \\
 & & InternVL\_2.5-8B & 80.79 & 76.63 (-4.16)  & 75.84 (-4.95)  & 84.36 (+3.57) & \textbf{8.52} \\ \midrule

 %----------------------------------------------------

 \multirow{28}{*}{With} &\multirow{4}{*}{Aircraft} & LLaVA-v1.5-7B & 19.43 & 21.71 (+2.28)  & 12.86 (-6.57)  & 54.57 (+35.14) & \textbf{41.71}\\
 & & LLaVA-v1.5-13B & 21.43 & 24.00 (+2.57)  & 10.57 (-10.86) & 75.14 (+53.71) & 64.57 \\
 & & Qwen2.5-VL-7B & \textbf{70.29} & \textbf{75.14} (+4.85)  & \textbf{45.43} (-24.86) & \textbf{96.00} (+25.71) & 50.57 \\
 & & InternVL\_2.5-8B & 54.29 & 56.57 (+2.28)  & 31.43 (-22.86) & 81.14 (+26.85) & 49.71 \\ \cmidrule(lr){2-8}

 &\multirow{6}{*}{Cars} & LLaVA-v1.5-7B & 25.82  & 32.35 (+6.53)  & 19.18 (-6.64)  & 79.08 (+53.26) & 59.90 \\
 & & LLaVA-v1.5-13B & 31.84 & 40.92 (+9.08)  & 26.84 (-5.00)  & 80.41 (+48.57) & 53.57 \\
 & & Qwen2.5-VL-3B & 76.73 & 75.41 (-1.32)  & 62.76 (-13.97) & 89.59 (+12.86) & 26.83 \\
 & & Qwen2.5-VL-7B & 82.24 & \textbf{84.49} (+2.25)  & 67.14 (-15.10) & \textbf{94.90} (+12.66) & 27.76 \\
 & & Qwen2.5-VL-32B & \textbf{85.10} & 82.96 (-2.14)  & \textbf{74.39} (-10.71) & 91.73 (+6.63) & \textbf{17.34} \\
 & & InternVL\_2.5-8B & 56.12 & 57.04 (+0.92)  & 44.80 (-11.32) & 75.51 (+19.39) & 30.71 \\ \cmidrule(lr){2-8}

 &\multirow{7}{*}{CUB} & LLaVA-v1.5-7B & 27.50 & 28.70 (+1.20)  & 23.40 (-4.10)  & 63.50 (+36.00) & 40.10 \\
 & & LLaVA-v1.5-13B & 28.30 & 33.70 (+5.40)  & 19.10 (-9.20)  & 76.40 (+48.10) & 57.30 \\
 & & Qwen2.5-VL-3B & 67.50 & 70.90 (+3.40)  & 52.50 (-15.00) & 87.30 (+19.80) & 34.80 \\
 & & Qwen2.5-VL-7B & 76.10 & 74.30 (-1.80)  & 50.70 (-25.40) & \textbf{93.90} (+17.80) & 43.20 \\
 & & Qwen2.5-VL-32B & 76.40 & 73.60 (-2.80)  & 61.00 (-15.40) & 88.60 (+12.20) & 27.60 \\
 & & Qwen3-VL-30B   & \textbf{77.80}	& \textbf{75.20} (-2.60)	& \textbf{61.90} (-15.90) & 88.50 (+10.70) & \textbf{26.60} \\
 & & InternVL\_2.5-8B & 61.40 & 60.00 (-1.40)  & 39.20 (-22.20) & 88.30 (+26.90) & 49.10 \\ \cmidrule(lr){2-8}

 &\multirow{5}{*}{Dogs} & LLaVA-v1.5-7B & 26.67 & 31.83 (+5.16)  & 16.00 (-10.67) & 76.67 (+50.00) & 60.67 \\
 & & LLaVA-v1.5-13B & 30.83 & 37.67 (+6.84)  & 20.00 (-10.83) & 76.83 (+46.00) & 56.83 \\
 & & Qwen2.5-VL-7B & 76.33 & \textbf{78.00} (+1.67)  & 58.83 (-17.50) & \textbf{93.33} (+17.00) & 34.50 \\
 & & Qwen3-VL-30B   & \textbf{77.67}	 & 77.67 (+0.00)	 & \textbf{67.67} (-10.00)	 & 88.50 (+10.83) & \textbf{20.83} \\
 & & InternVL\_2.5-8B & 59.33 & 57.67 (-1.66)  & 43.67 (-15.66) & 79.33 (+20.00) & 35.66 \\ \cmidrule(lr){2-8}

 &\multirow{4}{*}{Food} & LLaVA-v1.5-7B & 30.10& 58.22 (+28.12) & 32.87 (+2.77)  & 89.90 (+59.80) & 57.03 \\
 & & LLaVA-v1.5-13B & 58.42 & 71.88 (+13.46) & 56.63 (-1.79)  & \textbf{96.44 }(+38.02) & 39.81 \\
 & & Qwen2.5-VL-7B & \textbf{88.12} & \textbf{89.31} (+1.19)  & \textbf{84.36} (-3.76)  & 93.66 (+5.54) & 9.30 \\
 & & InternVL\_2.5-8B & 85.74 & 83.76 (-1.98)  & 81.39 (-4.35)  & 86.34 (+0.60) & \textbf{4.95} \\

\bottomrule
\end{tabular}
\caption{Accuracy (\%) of LVLMs across five datasets under different option-length conditions, with and without class names in answer choices. Results show LVLM performance on "Equal Long," "Equal Short," "Shorter Correct," and "Longer Correct" variants. Values in parentheses indicate differences from the "Equal Long" baseline for each model-dataset pair. $\Delta$ (Longer Correct - Shorter Correct) quantifies length bias magnitude. The best result for each setting and dataset is highlighted in bold.}
\label{tab:model_length_bias_comparison}
\vspace{-0.5cm}
\end{table*}

%%%%%%%%%%%%%%%%%%%%%%%%%%%%%%%%%%%%%%%%%%%%%%%%%%%%%%%

\subsection{Results for the Abstention (NoA) Settings}
\label{NoA_results}

\begin{figure}[!ht]
  \centering
  \vspace{-1em}
  \includegraphics[width=\linewidth]{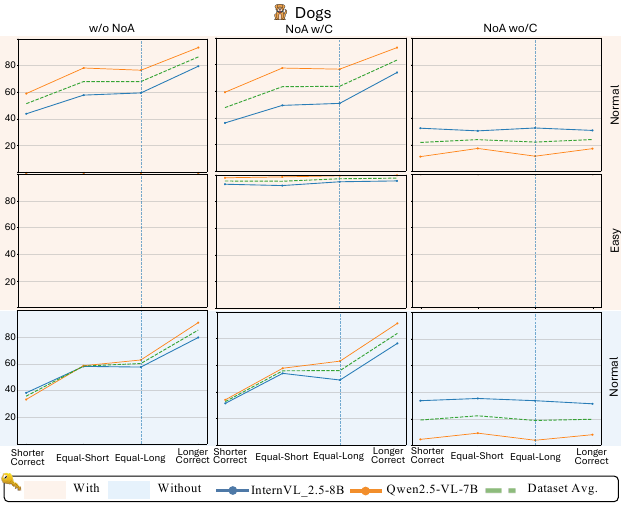}
   %\vspace{-3.5em}
   \caption{Accuracy (\%) of LVLMs on "None of the Above" (NoA) evaluation settings across Dogs datasets, reported both with and without class names in the answer options. "w/o NoA" denotes the original MCQ format without a NoA option. "NoA w/C" appends a fifth option, "E. None of the above", while keeping the correct answer present among A–D (over-abstention test). "NoA wo/C" removes the correct answer and uses "D. None of the above." as the only valid choice (forced-abstention test).}
   \label{fig:table_2_noa}
   %\vspace{-1em}
\end{figure}

Figure~\ref{fig:table_2_noa} and Table~\ref{tab:NoA_comparison} summarize LVLM performance under different abstention settings. We observe several consistent patterns.

\noindent\textbf{Without Class Names, Abstention Performance Collapses Completely.}
Without class names, LVLMs fail almost completely under abstention tests. Qwen2.5-VL-7B drops to 3.2\% on CUB and 3.83\% on Dogs under NoA wo/C, while InternVL\_2.5-8B falls to 17.7\% and 33.5\%, respectively. Such near-random behavior suggests that without semantic anchors, models cannot recognize when no valid answer is available and instead produce arbitrary guesses. This underscores the importance of class names as grounding cues for abstention, not just classification.

\noindent\textbf{Abstention Reveals Critical Robustness Failures in LVLMs.}
While LVLMs handle abstention reliably in trivial cases, they perform poorly under challenging conditions. They exhibit both false abstention (choosing NoA when the correct answer is present) and failure to abstain (guessing when abstention is correct). This highlights abstention as a critical but underexplored aspect of LVLM robustness with direct implications for safe deployment in real-world applications.

\begin{table*}[!ht]
\centering
\small
\begin{tabular}{lllccccc}
\toprule
\textbf{\makecell{Class\\Name}} &\textbf{Dataset} & \textbf{Setting} & \textbf{Model} & \textbf{\makecell{Equal\\Long}} & \textbf{\makecell{Equal\\Short}} & \textbf{\makecell{Shorter\\Correct}} & \textbf{\makecell{Longer\\Correct}} \\
\midrule

 \multirow{15}{*}{Without} &\multirow{7}{*}{CUB} &\multirow{2}{*}{w/o NoA} & Qwen2.5-VL-7B & 64.90 & 61.00 & 26.60 & 93.40 \\
 & & & InternVL\_2.5-8B & 62.50 & 57.10 & 37.80 & 88.30 \\ \cmidrule{3-8}
 
 & &\multirow{2}{*}{NoA w/C} & Qwen2.5-VL-7B & 64.10 & 60.20 & 28.10 & 93.30 \\
 & & & InternVL\_2.5-8B & 59.30 & 54.40 & 34.70 & 87.40 \\ \cmidrule{3-8}

  & &\multirow{2}{*}{NoA wo/C} & Qwen2.5-VL-7B & 3.20 & 5.80 & 3.20 & 5.40 \\
 & & & InternVL\_2.5-8B & 17.70 & 21.00 & 18.00 & 20.00 \\ \cmidrule{2-8}

 &\multirow{7}{*}{Dogs} &\multirow{2}{*}{w/o NoA} & Qwen2.5-VL-7B & 62.83 & 58.50 & 33.17 & 90.83 \\
 & & & InternVL\_2.5-8B & 57.50 & 58.00 & 38.17 & 79.67 \\ \cmidrule{3-8}

  & &\multirow{2}{*}{NoA w/C} & Qwen2.5-VL-7B & 63.00 & 57.67 & 34.00 & 91.33 \\
 & & & InternVL\_2.5-8B & 48.83 & 53.83 & 31.33 & 76.33 \\ \cmidrule{3-8}

  & &\multirow{2}{*}{NoA wo/C} & Qwen2.5-VL-7B & 3.83 & 9.17 & 4.50 & 8.00 \\
 & & & InternVL\_2.5-8B & 33.50 & 35.17 & 33.50 & 31.17 \\

 \midrule

 %----------------------------------------------------

 \multirow{22}{*}{With} &\multirow{7}{*}{CUB} &\multirow{2}{*}{w/o NoA} & Qwen2.5-VL-7B & 76.10 & 74.30 & 50.70 & 93.90 \\
 & & & InternVL\_2.5-8B & 61.40 & 60.00 & 39.20 & 88.30 \\ \cmidrule{3-8}

  & &\multirow{2}{*}{NoA w/C} & Qwen2.5-VL-7B & 75.00 & 73.20 & 50.70 & 93.50 \\
 & & & InternVL\_2.5-8B & 55.70 & 54.30 & 33.90 & 86.00 \\ \cmidrule{3-8}

  & &\multirow{2}{*}{NoA wo/C} & Qwen2.5-VL-7B & 20.00 & 20.50 & 12.20 & 19.70 \\
 & & & InternVL\_2.5-8B & 23.00 & 29.70 & 23.10 & 30.10 \\ \cmidrule{2-8}

 &\multirow{7}{*}{Dogs} &\multirow{2}{*}{w/o NoA} & Qwen2.5-VL-7B & 76.33 & 78.00 & 58.83 & 93.33 \\
 & & & InternVL\_2.5-8B & 59.33 & 57.67 & 43.67 & 79.33 \\ \cmidrule{3-8}

   & &\multirow{2}{*}{NoA w/C} & Qwen2.5-VL-7B & 76.33 & 77.17 & 59.00 & 92.50 \\
 & & & InternVL\_2.5-8B & 50.67 & 49.17 & 36.00 & 73.83 \\ \cmidrule{3-8}

  & &\multirow{2}{*}{NoA wo/C} & Qwen2.5-VL-7B & 11.33 & 17.17 & 11.00 & 17.00 \\
 & & & InternVL\_2.5-8B & 32.50 & 30.33 & 32.33 & 30.67 \\ \cmidrule{2-8}

  &\multirow{7}{*}{\makecell{Dogs\\(Easy)}} &\multirow{2}{*}{w/o NoA} & Qwen2.5-VL-7B & 100.0 & 100.0 & 100.0 & 100.0 \\
 & & & InternVL\_2.5-8B & 100.0 & 100.0 & 100.0 & 100.0 \\ \cmidrule{3-8}

   & &\multirow{2}{*}{NoA w/C} & Qwen2.5-VL-7B & 99.17 & 98.33 & 97.67 & 99.50 \\
 & & & InternVL\_2.5-8B & 94.67 & 91.83 & 92.83 & 95.33 \\ \cmidrule{3-8}

  & &\multirow{2}{*}{NoA wo/C} & Qwen2.5-VL-7B & 100.0 & 100.0 & 100.0 & 100.0 \\
 & & & InternVL\_2.5-8B & 100.0 & 100.0  & 100.0 & 100.0 \\ 

\bottomrule
\end{tabular}
%\vspace{-0.5cm}
\caption{Accuracy (\%) of LVLMs on "None of the Above" (NoA) evaluation settings across CUB and Dogs datasets, reported both with and without class names in the answer options. "w/o NoA" denotes the original MCQ format without a NoA option. "NoA w/C" appends a fifth option, "E. None of the above", while keeping the correct answer present among A--D (over-abstention test). "NoA wo/C" removes the correct answer and uses "D. None of the above." as the only valid choice (forced-abstention test). In the "NoA wo/C" setting, removing the correct answer from "Shorter Correct" leaves only long distractors, making its option pool structurally similar to "Equal Long"; likewise, removing the correct answer from "Longer Correct" leaves only short distractors, similar to "Equal Short". Accordingly, LVLM performance under "NoA wo/C" for these two conditions closely aligns with, though does not exactly match, "Equal Long" and "Equal Short", respectively.}
\label{tab:NoA_comparison}
\vspace{-0.5cm}
\end{table*}
%%%%%%%%%%%%%%%%%%%%%%%%%%%%%%%%%%%%%%%%%%%%%%%%%%%%%%%%%%%%%%%%

\subsection{Position Bias and Answer Distribution Analysis}
\label{sec:position_bias}

To ensure that our observed length bias is not confounded by position bias, we provide detailed analysis of answer position distributions and our experimental design controls.

\subsubsection{Answer Position Distribution Across Datasets}

Our length-controlled benchmarks inherit the answer position distributions from the underlying fine-grained MCQA datasets of \citet{atabuzzaman2025benchmarking}. Table~\ref{tab:position_distribution} presents the distribution of correct answers across positions A, B, C, and D for each dataset. While distributions show some natural variation, no position exhibits systematic over-representation that could confound our length manipulation experiments.

\begin{table}[!ht]
\centering
\small
\begin{tabular}{lcccc}
\toprule
\textbf{Dataset} & \textbf{A} & \textbf{B} & \textbf{C} & \textbf{D} \\
\midrule
Cars & 245 & 245 & 240 & 250 \\
CUB & 260 & 215 & 250 & 275 \\
Dogs & 170 & 145 & 160 & 125 \\
Food & 140 & 125 & 100 & 140 \\
\bottomrule
\end{tabular}
\caption{Distribution of correct answers across positions A, B, C, and D for each dataset. Distributions are inherited from base benchmarks and remain unchanged across all length-controlled variants.}
\label{tab:position_distribution}
\end{table}

\subsubsection{Position-Controlled Length Manipulation}

Critically, our length-controlled variant design explicitly neutralizes position effects. For every question, the correct answer remains in the \textit{same position} (A, B, C, or D) across all four variants: Equal Long, Equal Short, Shorter Correct, and Longer Correct. 

For example, consider a question where the correct answer originally appears at position C:
\begin{itemize}[noitemsep]
    \item \textbf{Equal Long:} Position C contains a detailed correct answer (all options long)
    \item \textbf{Equal Short:} Position C contains a brief correct answer (all options short)
    \item \textbf{Shorter Correct:} Position C contains a \textit{short} correct answer (distractors long)
    \item \textbf{Longer Correct:} Position C contains a \textit{long} correct answer (distractors short)
\end{itemize}

\begin{table}[!ht]
\centering
\small 
\begin{tabular}{llccc}
\toprule
\textbf{Model} & \textbf{Dataset} & \textbf{\makecell{Class \\name \\only}} & \textbf{\makecell{Shorter\\Correct}} & \textbf{\makecell{Longer\\Correct}} \\
\midrule
InternVL-2.5-8B & Dogs & 54.17 & 45.33 & 79.17 \\
InternVL-2.5-8B & CUB & 47.10 & 48.60 & 89.60 \\
Qwen2.5-VL-7B & CUB & 67.00 & 53.30 & 93.20 \\
Qwen2.5-VL-7B & Dogs & 75.83 & 60.17 & 93.33 \\
\bottomrule
\end{tabular}
\caption{Accuracy (\%) under class-name-only ablations. All conditions use identical answer positions, varying only the length of the correct option versus distractors.}
\label{tab:ablation_classname}
\end{table}

Because position is held constant while answer length is systematically varied, any performance differences across these variants cannot be attributed to position bias. Length is the only manipulated factor. This within-question design ensures that our results isolate length effects from positional preferences.

\subsubsection{Class-Name-Only Ablation}

To further verify that content variation does not confound length effects, we conducted additional ablations where only class names differ between conditions. Table~\ref{tab:ablation_classname} presents results where the baseline (Class-name-only) provides all options as bare class names, the Shorter Correct condition adds detailed descriptions only to distractors, and the Longer Correct condition adds detailed descriptions only to the correct answer.

Results demonstrate consistent accuracy shifts driven solely by length manipulation. For instance, Qwen2.5-VL-7B on CUB drops from 67.0\% to 53.3\% when the correct answer is short and rises to 93.2\% when it is long—a 39.9 percentage point gap attributable entirely to length. These results confirm that position bias does not explain our findings.

%%%%%%%%%%%%%%%%%%%%%%%%%%%%%%%%%%%%%%%%%%%%%%%%%%%%%%%%%%%%%%%%%%%%%

\end{document}